\documentclass[10pt,journal,letterpaper,compsoc]{IEEEtran}

\usepackage{cs}
\usepackage{mathsymb,xcolor}

\usepackage[small,labelfont=bf]{caption}

\usepackage[nocompress]{cite}

\usepackage{amsmath,amscd}
\usepackage{graphicx}
\graphicspath{{./}{./Fig/}{./Figs/}}

\usepackage{algorithm2e}
\let\savedalgorithm\algorithm
\let\savedendalgorithm\endalgorithm
\newenvironment{algorithmic}{%
\savedalgorithm
}{%
\savedendalgorithm
}

\usepackage{algorithm}

\newtheorem{theorem}{Theorem}[section]

\newtheorem{definition}{Definition}[section]
\newtheorem{proposition}{Proposition}[section]
\newtheorem{lemma}{Lemma}[section]

\def\margin{\varrho}
\def\lse{\mathrm{lse}}
\def\logit{\mathrm{logit}}
\def\erf{\mathrm{erf}}
\def\var{\mathrm{var}}

\def\mosek{{\sc Mosek}\xspace}
\def\cplex{{\sc CPlex}\xspace}

\def\Fada{f_\mathrm{ab}}
\def\varrhomean{{\bar \varrho}}
\def\bxi{ {\boldsymbol \xi}}
\def\edge{r}

\def\dif{{\mathrm d}}

\def\arcgv{{Arc-Gv}\xspace}
\def\adaboost{{AdaBoost}\xspace}
\def\lpboost{{LPBoost}\xspace}
\def\logitboost{{LogitBoost}\xspace}

\def\fig{{Fig.}}
\def\eq{{Equ.}}

\def\opt#1{\mathrm{Opt}_{(#1)}^\star}

\def\b#1{{\bf #1}}

\begin{document}

\title{
       On the Dual Formulation of Boosting Algorithms
}

\author{
  Chunhua Shen, ~~~  Hanxi Li
\thanks{%
C.~Shen and H.~Li are with NICTA (National ICT Australia), Canberra Research Laboratory, Canberra, ACT 2601, Australia;
  and  Australian
  National University, Canberra, ACT 0200, Australia. 
  E-mail: 
  \tt chunhua@me.com 
 }
 \thanks{%
 \color{blue}{%
Published     2010 Dec.; IEEE Transactions on Pattern Analysis and Machine Intellgience 32(12): 2216-31.
    }
 %
 }
\thanks{%
Digital Object Identifier: 10.1109/TPAMI.2010.47
 }
}

\markboth{IEEE TRANSACTIONS ON PATTERN ANALYSIS AND MACHINE
 INTELLIGENCE, VOL.~32, NO.~12, Dec.~2010}
 {SHEN AND LI: On the Dual Formulation of Boosting Algorithms}

\IEEEcompsoctitleabstractindextext{%

\begin{abstract}
      
      We study boosting algorithms from a new perspective.
      We show that the Lagrange dual problems of the  $ \ell_1$ norm regularized
      \adaboost,
      \logitboost and soft-margin \lpboost with generalized hinge loss 
      are all entropy maximization problems. 
      By looking at the dual problems of these boosting algorithms, we
      show that the success of boosting algorithms can be understood
      in terms of maintaining a better margin distribution by
      maximizing margins and at the same time controlling the margin
      variance. 
      We also theoretically prove that, approximately, the $ \ell_1$
      norm regularized \adaboost 
      maximizes the average margin, instead of the minimum margin. 
      The duality formulation also enables us to 
      develop column generation based optimization algorithms, which
      are totally corrective.  We show that they exhibit almost
      identical classification results to that of standard stage-wise additive
      boosting algorithms but  with much faster convergence rates.
      Therefore fewer weak classifiers are needed to build the
      ensemble using our proposed optimization technique.   

\end{abstract}



}

\maketitle

\thispagestyle{empty}

\section{Introduction}

 \IEEEPARstart{B}{oosting} 
         has attracted a lot of research interests since the
         first practical boosting algorithm, \adaboost, was introduced
         by Freund and Schapire \cite{Freund1997Decision}. 
         The machine learning community has spent much effort on
         understanding how the algorithm works
         \cite{Schapire1998Margin,Rudin2004Dynamics,Rudin2007Analysis}.
         However, up to date there are still questions about the
         success of boosting that are left unanswered 
         \cite{Wyner2008Evidence}. 
         In boosting, 
         one is given a set of training examples $ \bx_i \in
         \cX,
         i=1\cdots M $, with binary labels $ y_i $ being either $ +1 $
         or $ -1 $.  A boosting algorithm finds a convex linear 
         combination of weak classifiers (\aka base learners, weak
         hypotheses)
%
%
         that can achieve much better
         classification accuracy than an individual base classifier. 
         To do so, there are two unknown variables to be optimized.
         The first one is the base classifiers. An oracle is needed to
         produce base classifiers. The second one is the positive
         weights associated with each base classifier.

         \adaboost is
         one of the first and the most popular boosting algorithms for
         classification. Later, various boosting algorithms have been
         advocated.  For example, \logitboost 
         by Friedman \etal \cite{Friedman2000Additive}
         replaces \adaboost's exponential cost function with the
         function of logistic regression.
         MadaBoost \cite{Domingo2000MadaBoost}
         instead uses a modified exponential loss. 
         The authors of \cite{Friedman2000Additive} 
         consider boosting algorithms
         with a generalized additive model framework.  
         Schapire \etal \cite{Schapire1998Margin} showed 
         that \adaboost converges to a large margin solution. 
         However, recently it is pointed out that 
         \adaboost does not converge to the maximum margin solution
         \cite{Rudin2007Analysis,Ratsch2005Efficient}.
         Motivated by the success of the margin
         theory associated with support vector machines (SVMs),
         \lpboost was invented by 
         \cite{Grove1998Boosting,Demiriz2002LPBoost} with the
         intuition of maximizing the minimum margin of all training
         examples. The final optimization problem can be formulated as
         a linear program (LP). 
         It is observed that the hard-margin \lpboost does not perform
         well in most cases although it usually produces larger
         minimum margins. 
         More often \lpboost has worse generalization performance. 
         In other words, a higher minimum margin would not necessarily 
         imply a lower test error.
         Breiman \cite{Breiman1999Arc} also noticed the same
         phenomenon: his \arcgv algorithm has a minimum margin that
         provably converges to the optimal but \arcgv is inferior in
         terms of generalization capability. 
         Experiments on \lpboost and \arcgv have put the margin theory
         into serious doubt. 
         Until recently, Reyzin and Schapire \cite{Reyzin2006Boosting}
         re-ran Breiman's experiments by controlling weak classifiers'
         complexity. They found that 
         the minimum margin is indeed larger for \arcgv, but the overall
         margin distribution is typically better for \adaboost. 
         The conclusion is that the minimum margin is important, but
         not always at the expense of other factors.
         They also conjectured that maximizing the average margin,
         instead of the minimum margin, may result in better boosting
         algorithms. 
         Recent theoretical work
         \cite{Koltchinskii2002Empirical} 
         has shown the important role of the margin distribution  
         on bounding the generalization
         error of combined classifiers such as boosting and bagging.

         As the soft-margin SVM usually has a better classification
         accuracy than the hard-margin SVM, 
         the soft-margin \lpboost also performs better by relaxing 
         the constraints that all training examples must be correctly
         classified. Cross-validation is required to determine an
         optimal value for the soft-margin trade-off parameter.
         R\"atsch \etal \cite{Ratsch2002BoostSVM} showed the equivalence between SVMs
         and boosting-like algorithms. 
         Comprehensive overviews on boosting are given by
         \cite{Meir2003Boosting} and 
         \cite{Schapire2003Boosting}. 

         We show in this work that the Lagrange duals of
         $ \ell_1$ norm regularized
         \adaboost, \logitboost and \lpboost with generalized hinge
         loss are all  
         entropy maximization problems. 
         Previous work like    
         \cite{Collins2002Logistic,Kivinen1999Boosting,Lebanon2001Boosting}
         noticed the connection between boosting techniques and
         entropy maximization based on Bregman distances.   
         They did not  
         show that the duals of boosting algorithms are 
         actually entropy regularized \lpboost as
         we show in 
         \eqref{EQ:dual_ada0}, 
         \eqref{EQ:LPBoost10} and \eqref{EQ:dual_logit1}. 
         By knowing this duality equivalence, 
        we derive a general column generation (CG) based optimization
        framework that can be used to optimize arbitrary
        convex loss functions. In other words, we can easily design
        totally-corrective \adaboost, \logitboost and boosting 
        with generalized hinge loss, \etc.

         Our major contributions are the following: 
         \begin{enumerate}
            \item
                  We derive the Lagrangian duals of boosting
                  algorithms and show that most of them are entropy
                  maximization problems.
            \item
                  The authors of \cite{Reyzin2006Boosting}  conjectured
                  that                  
                  ``it may be fruitful to consider boosting algorithms
                  that greedily maximize the average or median margin
                  rather than the minimum one''.
                  We theoretically prove that, actually, $ \ell_1$ 
                  norm regularized \adaboost
                  approximately maximizes the average margin, instead
                  of the minimum margin. 
                  This is an important result in the sense that it
                  provides an alternative theoretical explanation
                  that is consistent with the margins theory and
                  agrees with the empirical observations made by
                  \cite{Reyzin2006Boosting}. 
            \item
                  We propose \adaboost-QP that directly optimizes the
                  asymptotic cost function of \adaboost. The
                  experiments confirm our theoretical analysis.
            \item
                  Furthermore, based on the duals we derive, we design
                  column generation based optimization techniques for
                  boosting learning. 
                  We show that the new algorithms have almost identical results
                  to that of standard stage-wise additive boosting
                  algorithms but with much faster convergence rates.
                  Therefore fewer weak classifiers are needed to build
                  the ensemble.   
         \end{enumerate}

         The following notation is used. 
         Typically, we use bold letters $\bu, \bv $ to denote vectors,
         as opposed to scalars $ u, v$ in lower case letters.
         We use capital letters $ U, V $ to denote matrices.
         All vectors are column vectors unless otherwise
         specified. 
         The inner product of two
         column vectors $ \bu$ and $ \bv$ are $ \bu^\T \bv = \sum_i
         u_i v_i$. Component-wise inequalities are expressed using 
         symbols $ \psd, \pd, \nsd, \nd $;
         \eg, $ \bu \psd \bv $ means for all the
         entries $ u_i \geq v_i $. $ \b{0} $ and $ \b{1} $ are column
         vectors with
         each entry being $ 0 $ and $ 1 $ respectively.
         The length will be clear from the context.  The abbreviation
         $ \sst $ means ``subject to''.
         We denote the domain of a function $ f(\cdot) $ as $ \dom f$.

         The paper is organized as follows. 
         Section \ref{sec:b1} briefly reviews several boosting
         algorithms for self-completeness. 
         Their corresponding duals are derived in Section
         \ref{sec:adaboost}. Our main results are also presented in
         Section
         \ref{sec:adaboost}. In Section \ref{sec:exp},
         we then present numerical experiments to
         illustrate various aspects of our new algorithms obtained in
         Section \ref{sec:adaboost}. We conclude the paper in the last
         section.

         \section{Boosting Algorithms}
         \label{sec:b1}

         We first review some basic ideas and the corresponding
         optimization problems of \adaboost, \lpboost and \logitboost,
         which are of interest in this present work.

         Let $ \cH $ be a class of base classifier 
         $ \cH = \{     
         h_j ( \cdot ):  \cX \rightarrow \Real, j = 1\cdots N  
         \}$. A boosting algorithm  seeks for a convex linear
         combination 
         \begin{equation}
               F( \bx  )   = 
               {\textstyle  \sum_{j = 1}^N w_j h_j (\bx) },
            \label{EQ:boost1}
         \end{equation}
         where $ \bw $ is the weak classifier weights to be optimized.
         \adaboost calls an oracle that selects a weak
         classifier $h_j(\cdot)$ at each iteration $ j $ and then
         calculates the weight $w_j$ associated with $h_j(\cdot)$.
         It is shown in \cite{Friedman2000Additive,Mason1999Boosting}
         that \adaboost (and many others like \logitboost)
         performs coordinate gradient descent  in function space, at
         each iteration choosing a weak classifier to include in the
         combination such that the cost function is maximally reduced.
         It is well known that coordinate descent has a slow
         convergence in many cases. From an optimization point of
         view, there is no particular reason to keep the weights 
         $ w_1,\cdots, w_{j-1} $ fixed at iteration $ j $. 
         Here we focus on the underlying mathematical programs that
         boosting algorithms minimize.

         \adaboost has proved to minimize the exponential loss
         function \cite{Collins2002Logistic}: 
         \begin{equation}
            \sminimize_{\bw}~   \sum_{i=1}^M \exp ( -y_i F(\bx_i) ),
            ~\sst ~\bw \psd \b{0}.    
            \label{EQ:AdaCost}
         \end{equation}
         Because the logarithmic function
         $ \log(\cdot)$ is a strictly monotonically increasing
         function, \adaboost equivalently solves  
         \begin{equation}
            \sminimize_{\bw}~
            \log \left( \sum_{i=1}^M 
                 \exp ( -y_i   F(\bx_i) )\right),~\sst
            ~\bw \psd \b{0}, \b{1}^\T \bw = \tfrac{1}{T}.
         \label{EQ:1}
         \end{equation}   
         Note that in the \adaboost algorithm, the 
         constraint $ \b{1}^\T \bw = \frac{1}{T}$ is not explicitly enforced.
         However, without this regularization 
         constraint, in the case of separable training data, 
         one can always make the cost function approach to 
         zero via enlarging the solution $\bw$ by an arbitrarily
         large factor.
         Here what matters is the sign of the classification
         evaluation function.
         Standard \adaboost seems to select the value of $ T $ by selecting how many
         iterations it runs. 
         Note that the relaxed
         version $ \b{1} ^ \T \bw \leq \tfrac{1}{T} $ is 
         actually equivalent to
         $ \b{1} ^ \T \bw = \tfrac{1}{T} $.\footnote{The
         reason why we do not write this constraint as 
                     $ \b{1}^\T \bw = T $ 
         will become clear later.}
         With the constraint  $ \b{1} ^ \T \bw \leq \tfrac{1}{T} $,
         if the final solution has
         $ \b{1} ^ \T \bw < \tfrac{1}{T} $, one can scale $ \bw $ such that
         $ \b{1} ^ \T \bw = \tfrac{1}{T} $ and clearly the scaled $ \bw $
         achieves a smaller loss. So the optimum
         must be achieved at the boundary.

         The boosting algorithm introduced in \eqref{EQ:1} is a
         $\ell_1$ norm regularized version of the
         original \adaboost because it is equivalent to 
         \begin{equation}
            \sminimize_{\bw}~
            \log \left( \sum_{i=1}^M 
                 \exp ( -y_i   F(\bx_i) )\right)
                 + \frac{1}{T'} \b{1}^\T \bw,~\sst
            ~\bw \psd \b{0}.
         \label{EQ:LagA}
         \end{equation}   
         For a certain $ T $, one can always find a $ T' $
         such that \eqref{EQ:1} and \eqref{EQ:LagA} have exactly the same 
         solution.
         Hereafter, we refer to this algorithm as AdaBoost$_{\ell1}$.


         We will show that it is very important to introduce this new
         cost function. All of our main results on AdaBoost$_{\ell1}$ 
         are obtained by
         analyzing this logarithmic cost function, not the
         original cost function. 
         Let us define the matrix $ H \in \Integer^{M \times N}$,
         which contains all the possible predictions of the training
         data using weak classifiers from the pool $ \cH $. 
         Explicitly $ H_{ij} = h_j ( \bx_i ) $ is the label
         ($\{+1,-1\} $) given by weak classifier $ h_j (\cdot) $ on
         the training example $ \bx_i $. 
         We use $ H_i = [ H_{i1} ~ H_{i2}\cdots H_{iN} ] $ to denote $
         i$-th row of $ H $, which constitutes the output of all the
         weak classifiers on the training example $ \bx_i $.
         The cost function of AdaBoost$_{\ell1}$ writes:
         \begin{equation}
            \sminimize_{\bw}~
         \log \left( \sum_{i=1}^M \exp ( -y_i  H_i \bw   )\right),~\sst
         ~\bw \psd \b{0}, \b{1}^\T \bw = \tfrac{1}{T}.
         \label{EQ:2}   
         \end{equation}
            We can also write the above program into
            \begin{equation}
               \sminimize_{\bw}~
               \log \left( \sum_{i=1}^M 
               \exp \left( 
                         - \frac{ y_i  H_i \bw }{ T }
                    \right)
                    \right),~\sst
               ~\bw \psd \b{0}, \b{1}^\T \bw = 1,
               \label{EQ:2A}
            \end{equation}
            which is exactly the same as \eqref{EQ:2}.
            In \cite{Rudin2007Analysis} the smooth margin that is
            similar but different to the logarithmic cost function,
            is used to analyze \adaboost's convergence
            behavior.\footnote{The smooth margin 
             in \cite{Rudin2007Analysis}
            is defined 
            as 
            \[
            \frac{   -  \log \bigl( \sum_{i=1}^M 
               \exp \left( 
                         -  y_i  H_i \bw 
                    \right)
                    \bigr)
                    }{
                    \b{1}^\T \bw
                    }.
                    \] 
           }

            Problem \eqref{EQ:2} (or \eqref{EQ:2A})
            is a convex problem in $ \bw $.  
            We know that the log-sum-exp function 
             $ \lse ( \bx ) =  \log(  \sum_{i=1}^M \exp x_i  ) $
            is
            convex \cite{Boyd2004Convex}.
            Composition with an affine mapping preserves
            convexity.
            Therefore, the cost function is convex. The constraints are
            linear hence convex too.    
       For completeness, we include the description of the standard
       stage-wise \adaboost and \arcgv in Algorithm~\ref{alg:Ada}.
       The only difference of these two algorithms is the way to
       calculate $ w_j $ (step (2) of Algorithm~\ref{alg:Ada}).
       For AdaBoost:
       \begin{equation}
            w_j = \frac{1}{2}  \log \frac{1+ \edge_j }{1 - \edge_j },   
          \label{EQ:ada_wj}
       \end{equation}
       where $ \edge_j $ is the edge of the weak classifier $ h_j (
       \cdot )$ defined as 
       $ 
       \edge_j = \sum_{i=1}^M  u_i y_i h_j ( \bx_i ) = 
       \sum_{i=1}^M  u_i y_i H_{ij}.
       $ 
       \arcgv modifies 
        \eqref{EQ:ada_wj} in order to maximize the
       minimum margin: 
       \begin{equation}
            w_j = \frac{1}{2}  \log \frac{1+ \edge_j }{1 - \edge_j } 
            - \frac{1}{2} \log \frac{1+ \margin_j }{1 - \margin_j }, 
          \label{EQ:arc_wj}
       \end{equation}
       where $ \margin_j $ is the minimum margin over all training
       examples of the combined classifier up to the current round:
       $
       \margin_j = \sminimize_i \{ y_i {\sum_{s=1}^{j-1}{w_s h_s(\bx_i) }
       }/{ \sum_{s=1}^{j-1}{w_s} } \}, 
       $
       with $ \margin_1 = 0$. 
       \arcgv clips $ w_j$ into $[0,1]$ by setting $ w_j = 1 $ if $
       w_j > 1$ and $ w_j = 0 $ if $ w_j < 0 $ \cite{Breiman1999Arc}.
       Other work such as \cite{Ratsch2005Efficient,Freund1999Adaptive}
       has used different approaches to determine
       $ \margin_j $ in \eqref{EQ:arc_wj}.

       %
       %
       %

   %
   %
   \SetVline

   \linesnumbered
   \begin{algorithm}[t]
     \caption{Stage-wise \adaboost, and \arcgv.}   
   \begin{algorithmic}
   \KwIn{Training set  $(\bx_i, y_i), y_i =\{ +1,-1\}, i = 1\cdots M$;
         maximum iteration
         $N_\mathrm{max}$.
   }
        { {\bf Initialization}:
            $ u^0_i = \frac{1}{ M }$, $ \forany i=1\cdots M$.
         }

   \For{ $ j = 1,\cdots, N_\mathrm{max}$ }
   {
      \begin{enumerate}
      \item
         Find a new base $ h_j(\cdot) $ using the distribution $ \bu^j 
         $\;
     \item
         Choose $ w_j $\;
     \item
        Update $ \bu $:
        $ u^{j+1}_i \propto u^{j}_i  
                                  \exp
                                  \left( 
                                    - y_i w_j h_j( \bx_i ) 
                              \right) $,
        $ \forany i$; and normalize $ \bu^{j+1}
                              $.
      
     \end{enumerate}
   }
   \KwOut{
         The learned classifier 
         $ F ( \bx ) = \sum_{j=1}^{N} w_j h_j( \bx ) $.
   }
   \end{algorithmic}
   \label{alg:Ada}
   \end{algorithm}

       %
       %


       \section{Lagrange Dual of Boosting Algor\-i\-t\-h\-ms}
       \label{sec:adaboost}


      Our main derivations are based on a form of duality termed 
      convex conjugate or Fenchel duality.

         \begin{definition}(Convex conjugate)         
         Let $f:\Real^n \rightarrow \Real$.
         The function $f^*:\Real^n \rightarrow \Real$,
         defined as 
         \begin{equation}
                     f^*( \bu ) =
                     \sup_{ \bx \in \dom f } 
                         \left( \bu^\T \bx - f(\bx)  
                      \right),    
            \label{EQ:conjugate}
         \end{equation}
         is called the convex conjugate 
         (or Fenchel duality) of
         the function $ f(\cdot) $. The domain of the conjugate
         function consists of $ \bu \in \Real^n  $
         for which the supremum is finite.
         \end{definition}
         $ f^*(\cdot) $ is always a convex function because it is
         the pointwise supremum of a family of affine functions of $
         \bu $. This is true even if $f(\cdot)$ is non-convex
         \cite{Boyd2004Convex}.

         \begin{proposition}(Conjugate of log-sum-exp)
            The conjugate of the log-sum-exp function
            is the negative
            entropy function, restricted to the probability simplex.
            Formally, for $ \lse ( \bx ) =  \log(  \sum_{i=1}^M \exp x_i  ) $,
            its conjugate is:
                       \[ 
                           \lse^* ( \bu ) = 
                           \begin{cases}
                              \sum_{i=1}^M u_i \log u_i, 
                                 &\text{if $ \bu \psd \b{0}$ and $\b{1}^\T \bu = 1 $;}\\
                              \infty 
                                   &\text{otherwise.} 
                           \end{cases}
                       \]
                       We interpret $0 \log 0$ as $0$.
         \label{prop:1}
         \end{proposition}
         Chapter~$3.3$ of \cite{Boyd2004Convex} gives this result. 
         \begin{theorem}
            \label{thm:1}
            The dual of  AdaBoost$_{\ell1}$ is a Shannon entropy maximization
            problem,
            which writes,
             \begin{align}
              \label{EQ:dual_ada0}
            \smaximize_{r, \bu} ~ &
            \frac{r}{T} - {\textstyle \sum_{i=1}^M}{ u_i \log u_i } \notag \\
            \sst ~ & {\textstyle \sum_{i=1}^M} { u_i y_i H_i }
                                          \nsd - r \b{1}^\T, \\
                   & \bu \psd \b{0}, \b{1}^\T \bu = 1. \notag
         \end{align}
         \end{theorem}
      \begin{proof}
      %
      %
      To derive a Lagrange dual of AdaBoost$_{\ell1 } $,    
         we first introduce a new
         variable $ \bz \in \Real^M $ such that its $i$-th entry $ z_i
         = - y_i H_i \bw $, to obtain the equivalent problem
         \begin{align}
            \label{EQ:2B}
              \sminimize_{\bw} ~&
              \log \left( 
              {\textstyle \sum_{i=1}^M \exp z_i }
              \right)  \notag \\
           \sst ~&
            z_i =  -y_i  H_i \bw ~ (\forany i=1,\cdots,M),  \\
          & \bw \psd \b{0}, \b{1}^\T \bw = \tfrac{1}{ T }. \notag
         \end{align}
         The Lagrangian $L(\cdot)$ associated
         with the problem \eqref{EQ:2} is
         \begin{align}
            \label{EQ:Lag1}
         L ( \bw, \bz, \bu, \bq, r )   
           & = \log \left( \sum_{i=1}^M \exp z_i \right)
            - \sum_{i=1}^M u_i (  z_i + y_i H_i \bw  )
           \notag \\ 
           & - \bq^\T
           \bw - r ( \b{1} ^ \T \bw - \tfrac{1}{T} ),
         \end{align}
         with $ \bq \psd 0 $. 
         The dual function is
         \begin{align}
            \inf_{ \bz, \bw } L 
                      &=
            \inf_{ \bz, \bw } \log \left( \sum_{i=1}^M \exp z_i
            \right)   -   \sum_{i=1}^M u_i z_i 
                      + \frac{r}{T} \notag \\
                      &- 
                      \overbrace{
                      \left(  \sum_{i=1}^M u_i y_i H_i    
                       + \bq^\T  + r \b{1}^\T 
                          \right)
                          }^{\text{ must be $\boldsymbol 0$}}
                          \bw   
                      \notag     \\ 
                      &= \inf_{ \bz }
                       \log \left( \sum_{i=1}^M \exp z_i
                            \right)
                       - \bu^\T \bz + \frac{r}{T}  \notag   \\
                       &=
                       \overbrace{
                           - \sup_{\bz} 
                       \left[
                           \bu ^\T \bz 
                       - \log \left( \sum_{i=1}^M \exp z_i 
                       \right) 
                       \right]
                       }^ 
                       {- \lse^*(\bu)~(\text{see
                       Proposition~\ref{prop:1}})  } + \frac{r}{T}
                       \notag \\
                       &=
                       - \sum_{i=1}^M u_i \log u_i + \frac{r}{T}.
            \label{EQ:4}
         \end{align}
         By collecting all the constraints and eliminating $ \bq $,
         the dual of Problem \eqref{EQ:2} 
         is \eqref{EQ:dual_ada0}. 
      \end{proof}
      Keeping two variables $ \bw $ and $ \bz $, and
      introducing new equality constraints $  z_i = - y_i H_i \bw
      $, $ \forany i$, is essential to derive the above simple and
      elegant Lagrange dual.  Simple equivalent reformulations of a
      problem can lead to very different dual problems.  Without
      introducing new variables and equality constraints, one would
      not be able to obtain \eqref{EQ:dual_ada0}.
      Here we have considered the {\em negative margin} $ z_i $ to be
      the central objects of study.
      In \cite{Rifkin2007Value}, a similar idea has been used to
      derive different duals of kernel methods, which leads to
      the so-called {\em value regularization}. We focus on boosting algorithms
      instead of kernel methods in this work.
      Also note that we would have the following dual if we work directly
      on the cost function in \eqref{EQ:1}:
      \begin{align}
           \label{EQ:2F}
            \smaximize_{r, \bu} ~ &
            \frac{r}{T} - {\textstyle \sum_{i=1}^M}{ u_i \log u_i } +
            \b{1}^\T \bu \notag \\
            \sst ~ & {\textstyle \sum_{i=1}^M} { u_i y_i H_i }
                                          \nsd - r \b{1}^\T, 
                    \bu \psd \b{0}.
      \end{align}
      No normalization requirement $ \b{1} ^ \T \bu = 1 $ is imposed.
      Instead, $ \b{1} ^ \T \bu $ works as a regularization term. 
      The connection between \adaboost and \lpboost is not clear with
      this dual.

      Lagrange duality between problems \eqref{EQ:2} and
      \eqref{EQ:dual_ada0} assures  that weak duality and strong duality
      hold. 
      Weak duality says that any feasible solution of
      \eqref{EQ:dual_ada0} produces a lower bound of the original 
      problem \eqref{EQ:2}. Strong duality tells us the optimal value
      of \eqref{EQ:dual_ada0} is the same as the optimal value of
      \eqref{EQ:2}. The weak duality is guaranteed by the Lagrange
      duality theory. The strong duality holds since the primal
      problem \eqref{EQ:2} is a convex problem that satisfies Slater's
      condition \cite{Boyd2004Convex}.

      To show the connection with \lpboost, 
      we equivalently rewrite the above formulation by reversing
      the sign of $ r $ and multiplying the cost function with $ T
      $, $(T > 0)$:
      \begin{align}
         \label{EQ:dual_ada1}
         \sminimize_{r, \bu} ~ &
         r + T {\textstyle \sum_{i=1}^M}{ u_i \log u_i } \notag \\
         \sst ~ & {\textstyle \sum_{i=1}^M} { u_i y_i H_{ij} }
         \leq r 
         ~(\forany j=1,\cdots,N), \\
         & \bu \psd \b{0}, \b{1}^\T \bu = 1. \notag
         \end{align}

         Note that the constraint $ \bu \psd \b{0} $ is implicitly enforced
         by the logarithmic function and thus it can be dropped when
         one solves \eqref{EQ:dual_ada1}.
         %
         %
         %
         %
         %

\subsection{Connection between AdaBoost$_{\ell1}$ and Gibbs free energy}

      Gibbs free energy is the chemical potential that is minimized
      when a system reaches equilibrium at constant pressure and
      temperature. 

      Let us consider a system that has $ M $ states at temperature $
      T $. Each state has energy $ v_i $ and probability $ u_i $ of
      likelihood of occurring. The Gibbs free energy of this system
      is related with its average energy and entropy, namely:
      \begin{equation}
         G( \bv, \bu ) = \bu ^\T \bv + 
         T {\textstyle \sum_{i=1}^{M} u_i \log u_i}.
         \label{EQ:Gibbs}
      \end{equation}
      When the system reaches equilibrium, $ G(\bv,\bu)$ is minimized. 
      So we have 
      \begin{equation}
         \sminimize_\bu~G( \bv, \bu ),
                    ~\sst~\bu \psd \b{0}, \b{1}^\T \bu = 1. 
         \label{EQ:GibbsMin}
      \end{equation}
      The constraints ensure that $ \bu$ is a probability distribution.
      
      Now let us define vector  $ \bv_j $ with its entries 
      being $ v_{ij} = y_i H_{ij} $.  $ v_{ij} $ is the energy
      associated with state $ i $ for case $ j $. 
      $ v_{ij} $ can only take discrete binary values $ +1 $ or $ -1
      $.
      We 
      rewrite our dual optimization problem 
      \eqref{EQ:dual_ada1} into
      \begin{align}
         \sminimize_\bu~ & \overbrace{ 
                                 \smaximize_j \{ \bu^\T \bv_j \}
                                 }
                                 ^{\text{worst case energy vector}
                                 ~\bv_j }
                          + T \sum_{i=1}^M u_i \log u_i, 
                          \notag \\ 
                          ~\sst~&\bu \psd \b{0}, \b{1}^\T \bu=1.
         \label{EQ:Gibbs1}
      \end{align}
      This can be interpreted as 
      finding the minimum Gibbs free energy
      for the {\em worst} case energy vector.

\subsection{Connection between AdaBoost$_{\ell1}$ and \lpboost}

\label{sec:adalp}

            First let us recall the basic concepts of \lpboost.
            The idea of \lpboost is to maximize the minimum margin
            because it is believed that the minimum margin plays a
            critically important role in terms of generalization error
            \cite{Schapire1998Margin}. The hard-margin \lpboost
            \cite{Grove1998Boosting}
            can be formulated as  
            \begin{align}
                \label{EQ:LPBoost1}
                \smaximize_{\bw} ~   
                \overbrace{
                \sminimize_i \{ y_i H_i \bw \}}^{ {\text{minimum
                margin}} }, ~
                            \sst ~  \bw \psd \b{0}, \b{1}^\T \bw =1.  
            \end{align}
         This problem can be solved as an LP. Its dual is also an LP:
            \begin{align}
           \label{EQ:dual_LP1}
            \sminimize_{r, \bu} ~ \,
            r  \,\,\,\,
            \sst ~ & {\textstyle \sum_{i=1}^M} { u_i y_i H_{ij} }
                                          \leq r 
                                          ~(\forany j=1,\cdots,N), \notag \\
                   & \bu \psd \b{0}, \b{1}^\T \bu = 1. 
         \end{align}
         \arcgv has been shown asymptotically to a solution of 
         the above LPs \cite{Breiman1999Arc}. 
         
         The performance deteriorates when no linear combination of
         weak classifiers can be found that separates the training
         examples.  
         By introducing slack variables, we get the soft-margin LPBoost
         algorithm
            \begin{align}
                \label{EQ:SoftLPBoost1}
                \smaximize_{\bw, \varrho, \bxi} ~ &  
                \varrho - D \b{1}^\T \bxi  \notag \\
                \sst ~ &
                 y_i H_i \bw \geq \varrho - \xi_i, (\forany i
                 =1,\cdots,M),
                  \\
               &  \bw \psd \b{0}, \b{1}^\T \bw =1, \bxi \psd \b{0}. \notag  
            \end{align}
         Here $ D $ is a trade-off  parameter that controls the balance
         between training error and margin maximization. 
         The dual of \eqref{EQ:SoftLPBoost1} is similar to the
         hard-margin case except that the dual variable $ \bu $ is
         capped:
              \begin{align}
           \label{EQ:dual_SoftLP1}
            \sminimize_{r, \bu} ~ \,
            r  \,\,\,\, 
            \sst ~  & {\textstyle \sum_{i=1}^M} { u_i y_i H_{ij} }
                                          \leq r 
                                          ~(\forany j=1,\cdots,N), \notag \\
                   & D \b{1} \psd \bu \psd \b{0}, \b{1}^\T \bu = 1. 
         \end{align}

            Comparing \eqref{EQ:dual_ada1} with hard-margin \lpboost's
            dual,
            it is easy to see that the only difference is the entropy
            term in the cost function. If we set $ T = 0 $,
            \eqref{EQ:dual_ada1} reduces to the hard-margin \lpboost. 
            In this sense, we can view AdaBoost$_{\ell1}$'s dual as entropy
            regularized hard-margin \lpboost. 
            Since the regularization coefficient $ T $ is always
            positive, the effects of the entropy regularization term
            is to encourage the distribution $ \bu $ as uniform as
            possible (the negative entropy $ \sum_{i=1}^M u_i \log u_i
            $ is the Kullback-Leibler distance between $ \bu $ and the
            uniform distribution). 
            This may explain the underlying reason of \adaboost's
            success over hard-margin \lpboost: 
            To limit the weight distribution $ \bu $ leads to better
            generalization performance.  
            But, {\em why and how}?  
            We will discover the mechanism in Section 
            \ref{sec:sample_weight}.

            When the regularization coefficient, $ T $, is sufficiently
            large, the entropy term in the cost function dominates.  In
            this case, all discrete probability $ u_i $ become almost
            the same  and therefore gather around the center of the
            simplex $\{ \bu \psd \b{0}, \b{1}^\T \bu =1  \}$.  As $ T $
            decreases, the solution will gradually shift to the
            boundaries of the simplex to find the
            best mixture that best approximates the maximum. 
            Therefore, $ T $ can be also viewed as a homotopy
            parameter that bridges a maximum entropy problem with 
            uniform distribution $ u_i = 1/M$ ($ i = 1, ..., M$),
            to a solution of the max-min problem 
            \eqref{EQ:LPBoost1}.

            This observation is also consistent with the soft-margin
            \lpboost. We know that soft-margin \lpboost often
            outperforms
            hard-margin \lpboost
            \cite{Grove1998Boosting,Demiriz2002LPBoost}.
            In the primal, it is usually
            explained that the hinge loss of soft-margin is more
            appropriate for classification. 
            The introduction of slack variables in the primal actually
            results in box constraints on the weight distribution 
            in the dual.
            In other words the $\ell_\infty$ norm 
            of $ \bu $, $ \fnorm{ \infty}{ \bu }  $, is capped. 
            This capping
            mechanism is {\em harder} than the entropy regularization 
            mechanism of AdaBoost$_{\ell1 } $. Nevertheless, both are beneficial
            on inseparable data. 
            In \cite{Shalev-Shwartz2008Equ}, it is proved that 
            soft-margin \lpboost actually maximizes the average of $
            1/D $ smallest margins.

            Now let us take a look at the cost function of \adaboost
            and \lpboost in the primal. The log-sum-exp cost employed
            by \adaboost can be viewed as a smooth approximation of the
            maximum function because of the following inequality:
            \[
            \smaximize_i a_i \leq \log \bigl(  
                  {\textstyle \sum_{i=1}^M  \exp
                              a_i }
                        \bigr)  \leq \smaximize_i a_i + \log M.  
            \]
            Therefore, \lpboost uses a hard maximum (or minimum)
            function while \adaboost uses a soft approximation of the
            maximum (minimum) function.
%
            We try to explain why \adaboost's soft cost function is
            better than \lpboost's\footnote{Hereafter, 
            we use \lpboost to denote hard-margin \lpboost unless
            otherwise specified.
            }
            hard cost function next.

            \subsection{AdaBoost$_{\ell1}$ controls the margin variance via
            maximizing the entropy of the weights on the training
            examples}
            \label{sec:sample_weight}

            In \adaboost training, there are two sets of weights: the
            weights of the weak classifiers $ \bw $ and the weights on
            the training examples $ \bu $.  In the last section, we
            suppose that to limit $ \bu $
            is 
            beneficial for classification performance.   
            By looking at the Karush-Kuhn-Tucker (KKT) conditions of
            the convex program that we have formulated, we are able to
            reveal the relationship between the two sets of weights.
            More precisely, we show how  
            AdaBoost$_{\ell1}$ (and AdaBoost\footnote{ 
            We believe that the only difference between AdaBoost$_{\ell1}$ 
            and AdaBoost is on the optimization method employed by each algorithm.
            We conjecture that some theoretical results on AdaBoost$_{\ell1}$ derived 
            in this paper may also 
            apply to AdaBoost. 
            })
            controls the margin
            variance by optimizing the entropy of weights $ \bu $.

            Recall that we have to introduce new equalities $ z_i = -y_i H_i
            \bw, \forany i$ in order to obtain the dual
            \eqref{EQ:dual_ada0} (and \eqref{EQ:dual_ada1}). 
            Obviously $ z_i $ is the negative margin of sample 
            $ \bx_i $. Notice that the Lagrange multiplier 
            $ \bu $ is associated with these equalities. 
            Let $( \bw^\star $, $ \bz^{\star} )$
            and $( \bu^{\star}, \bq^{\star}, r^{\star} )$ be 
            any primal and dual optimal points with zero duality gap.
            One of the KKT conditions tells us 
            \begin{align}
               \grad_\bz 
               L( \bw^\star, \bz^{\star}, \bu^{\star}, \bq^{\star}, r^{\star} ) 
               = 0.
               \label{EQ:KKT1}
            \end{align}
            The Lagrangian $ L(\cdot ) $ is defined in
            \eqref{EQ:Lag1}.
            This equation follows 
            \begin{equation}
               \label{EQ:weight2}
                        u_i^\star = \frac{ \exp z_i^\star}
                                         {\sum_{i=1}^M \exp z_i^\star},
                        ~\forany
                        i=1,\cdots M. 
            \end{equation}
            \eq~\eqref{EQ:weight2} guarantees that
            $ \bu^\star $ is a probability
            distribution. Note that \eqref{EQ:weight2} is actually the
            same as the update rule used in \adaboost.
            The optimal value\footnote{Hereafter we use the
            symbol $\opt{\cdot} $ 
            to denote the optimal value of 
            Problem ($ \cdot$).}
            of the Lagrange dual problem 
            \eqref{EQ:dual_ada0}, which we denote 
            $\opt{\ref{EQ:dual_ada0}}$, 
            equals to the optimal value of
            the original problem \eqref{EQ:2} (and \eqref{EQ:2B})
            due to the strong
            duality, hence
            $ 
                  \opt{\ref{EQ:2}} = \opt{\ref{EQ:dual_ada0}} 
            $.

            From \eqref{EQ:weight2}, at optimality we  
            have 
            \begin{align}
               - z_i^\star & = - \log u_i^\star - \log 
                      \bigl(
                        \textstyle{  
                         \sum_{i=1}^M \exp z_i^\star 
                         }
                      \bigr)   \notag \\
                      & = - \log u_i^\star
                        - \opt{\ref{EQ:dual_ada0}}
                        \notag \\
                        & =- \log u_i^\star
                        - \opt{\ref{EQ:2}}, ~
                     \forany i=1,\cdots M.
               \label{EQ:weight3}
            \end{align}
            This equation suggests that, after convergence, 
            the margins' values are determined by the weights
            on the training examples $ \bu^\star $ 
            and the cost function's value. 
            From \eqref{EQ:weight3},  
            the margin's variance is
            entirely determined by $ \bu^\star $:
            \begin{equation}
               \var \{ - \bz^\star \} =
               \var \{  \log\bu^\star \} + 
               \var \{ \opt{\ref{EQ:2}} \}
                = \var \{  \log\bu^\star \}.
               \label{EQ:variance}
            \end{equation}

            We now understand the reason why capping $ \bu $ as
            \lpboost does, or uniforming $ \bu $ as \adaboost does can
            improve the classification performance. 
            These two equations reveal the important role that the 
            weight distribution $ \bu $ plays in \adaboost.
            All that we knew previously is that the weights on the
            training examples measure how difficult an individual
            example can be correctly classified. 
            In fact, besides that, 
            the weight distribution on the training examples 
            is also a {\em proxy} for 
            minimizing the margin's distribution divergence.
            From the viewpoint of optimization, this is an interesting
            finding. In \adaboost,
            one of the main purposes is to control the   
            divergence of the margin distribution, which may not be easy 
            to optimize {\em directly} because a margin can take a
            value out of the range $ [0, 1]$,
            where entropy is not applicable.   
            \adaboost's cost function allows one to do so {\em
            implicitly} in the primal but {\em explicitly} in the
            dual. 
           %
           %
            A future research topic is to apply this idea to other
            machine learning problems.

            The connection between the dual variable $ \bu $ and 
            margins tells us that \adaboost often seems to optimize the
            minimum margin (or average margin? We will answer
            this question in the next section.)
            but also it considers another quantity related to the 
            variance of the margins. In the dual
            problem~\eqref{EQ:dual_ada1}, minimizing the maximum edge
            on the weak classifiers contributes to maximizing the
            margin. At the same time, minimizing the negative entropy
            of weights on training examples contributes to controlling
            the margin's variance. 
            We make this useful observation by examining the dual
            problem as well as the KKT optimality conditions.  
            But it remains unclear about the exact statistics
            measures that \adaboost optimizes. 
            Next section presents a complete answer to this question
            through analyzing \adaboost's primal optimization problem.

\begin{figure}[t!]
                \begin{center}
                  \includegraphics[width=0.35\textwidth]{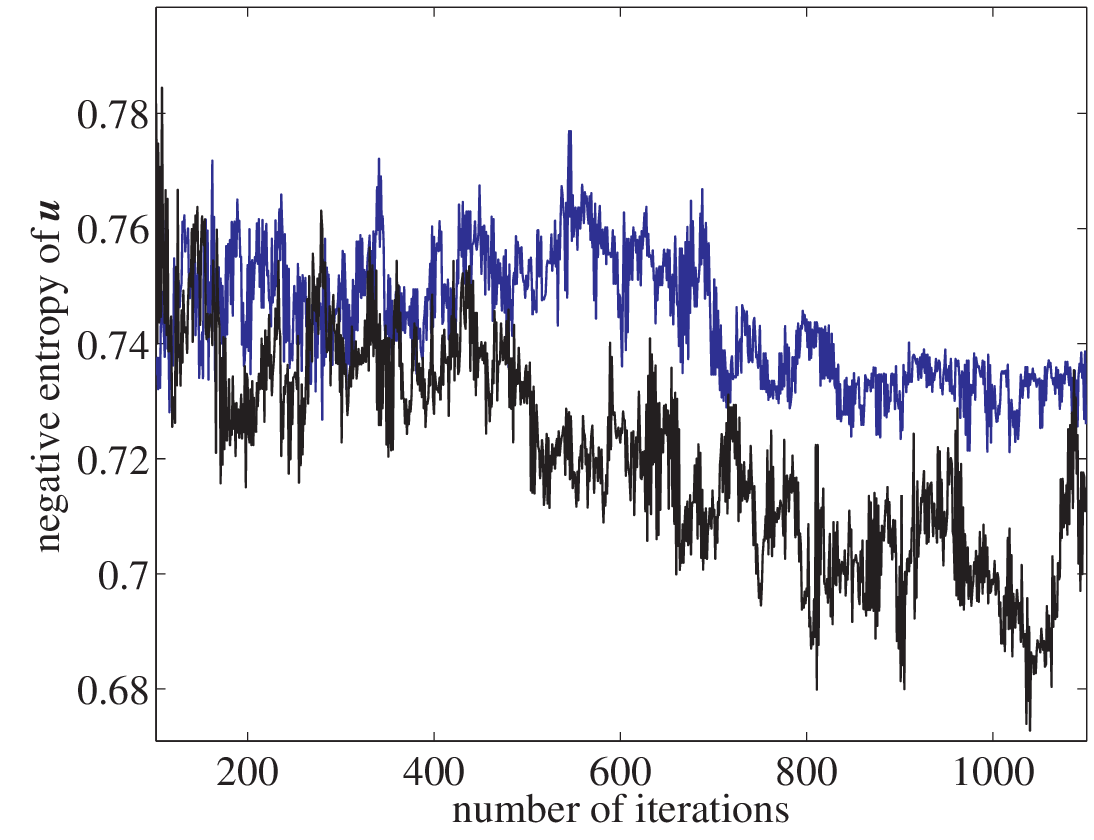}
                  \includegraphics[width=0.35\textwidth]{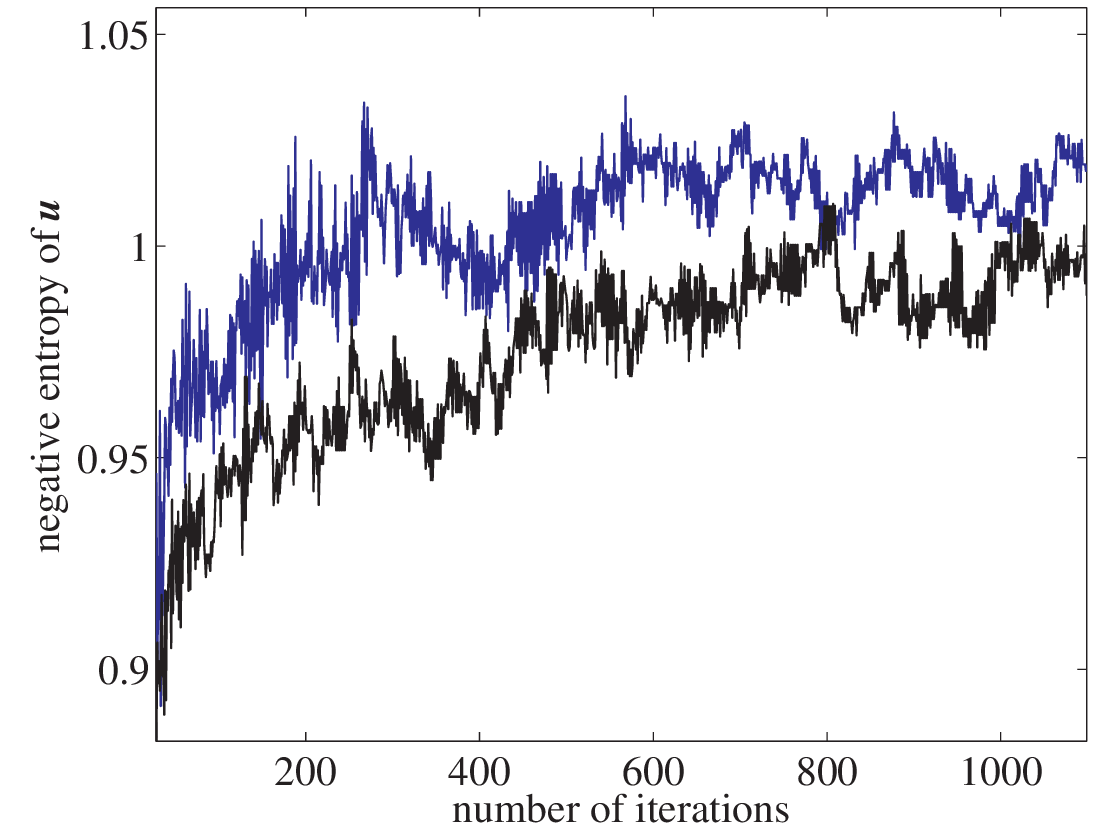}
                \end{center}
                \caption{Negative entropy of $ \bu $
                produced by the standard \adaboost and \arcgv at each iteration 
                 on datasets {\em breast-cancer} and {\em
                 australian}  respectively. The negative
                 entropy produced by \adaboost (black) is consistently lower
                 than the one by \arcgv (blue). 
                }
                \label{fig:entropy}
             \end{figure}

            We know that \arcgv chooses $ \bw$ in a different way from
            \adaboost. Therefore \arcgv optimizes a different cost
            function and does not minimize the negative entropy of $
            \bu $ any more. 
            We expect that \adaboost will have a more uniform
            distribution of $ \bu $. 
            We run \adaboost and \arcgv with decision stumps on two
            datasets {\em breast-cancer} and {\em
            australian} (all datasets used in this
            paper are available at \cite{LIBSVMdata2001} unless
            otherwise specified).
            \fig~\ref{fig:entropy} displays the results. \adaboost
            indeed has a small negative 
            entropy of $ \bu $ in both experiments, which agrees with 
            our prediction.

            It is evident now that AdaBoost$_{\ell1}$ 
            controls the variance of
            margins by regularizing the Shannon entropy of the
            corresponding dual variable $ \bu $. 
            For on-line learning algorithms there are two main
            families of regularization strategies: entropy
            regularization and regularization using squared Euclidean
            distance. A question that naturally arises here is:
            What happens if we use squared Euclidean distance to
            replace the entropy in the dual of AdaBoost$_{\ell1}$
            \eqref{EQ:dual_ada1}? In other words, {\em can we directly
            minimize the variance of the dual variable $ \bu $ to
            achieve the purpose of controlling the variance of
            margins}?
            We answer this question by having a look at the
            convex loss functions for classification. 
            
            \fig~\ref{fig:lossfun} plots four popular convex loss
            functions. 
            It is shown in \cite{Bartlett2004Convexity} 
            that as the data size increases, practically all popular
            convex loss functions are Bayes-consistent, although
            convergence rates and other measures of consistency may
            vary.
            In the context of boosting, \adaboost, \logitboost and
            soft-margin \lpboost use exponential loss, logistic loss
            and hinge loss respectively. 
            Here we are interested in the squared hinge loss. \logitboost
            will be discussed in the next section.  
            As mentioned, in theory, there is no particular reason to
            prefer hinge loss to squared hinge loss. 
            Now if squared hinge loss is adopted, 
            the cost function of 
            soft-margin \lpboost \eqref{EQ:SoftLPBoost1} becomes 
            $$
            \smaximize_{\bw, \margin, \bxi}~ \margin - D 
            \textstyle 
            \sum_{i=1}^M
            \xi_i^2,
            $$
            and the constraints remain the same as in
            \eqref{EQ:SoftLPBoost1}.  
            Its dual is easily derived\footnote{The primal constraint
            $ \bxi \psd 0 $ can be dropped because it is implicitly
            enforced.}            
  \begin{align}
           \label{EQ:dual_SoftDP1}
            \sminimize_{r, \bu} ~ &
            r  + \tfrac{1}{4D} {\textstyle \sum_{i=1}^M u_i^2 }\notag \\
            \sst ~ & {\textstyle \sum_{i=1}^M} { u_i y_i H_{ij} }
                                          \leq r 
                                          ~(\forany j=1,\cdots,N), \\
                   &  \bu \psd \b{0}, \b{1}^\T \bu = 1. \notag
         \end{align}       
      We can view the above optimization problem as variance
      regularized \lpboost. 
      In short, to minimize the variance of the dual variable $ \bu $ 
      for controlling the margin's variance, one can simply replace 
      soft-margin \lpboost's  hinge loss with the squared hinge loss. 
      Both the primal and dual problems are quadratic programs (QP) and
      hence can be efficiently solved using off-the-shelf QP solvers
      like \mosek \cite{Mosek}, \cplex \cite{Cplex}.

      Actually we can generalize the hinge loss into 
            \[ 
                  \left( \smaximize\{ 0, 1 - yF ( \bx ) \} \right)^p.    
            \]
      When $ p \geq 1$, the loss is convex. $ p = 1$ is the hinge loss
      and $ p = 2 $ is the squared hinge loss. 
      If we use a generalized hinge loss ($ p > 1 $) for boosting, we
      end up with a regularized \lpboost which has the format:
      \begin{equation}
         \sminimize_{r, \bu} ~ 
         r  +  D^{1-q} ( p ^ { 1-q} - p^{-q} )
                  {\textstyle \sum_{i=1}^M u_i^q },     
         \label{EQ:LPBoost10}
      \end{equation} 
      subject to the same constraints as in 
      \eqref{EQ:dual_SoftDP1}.
       Here
      $ p $ and $ q $ are dual to each other by 
      $ \frac{1}{p} + \frac{1}{q} = 1$. 
      It is interesting that \eqref{EQ:LPBoost10} can also be seen as
      entropy regularized \lpboost; more precisely, Tsallis entropy
      \cite{Tsallis1988Entropy}
      regularized \lpboost. 
      \begin{definition}(Tsallis entropy)         
         Tsallis entropy 
         is a generalization of the Shannon entropy,
         defined as 
         \begin{equation}
            S_q ( \bu ) = \frac{1 - \sum_i u_i^q } { q - 1}, ~~
            (\bu \psd 0, \b{1}^\T \bu = 1). 
            \label{EQ:Tsallis}
         \end{equation}
         where $ q $ is a real number. In the limit as $ q \rightarrow
         1 $, we have 
         $ u_i^{q-1} = \exp ( (q-1) \log u_i )  \simeq 1 + (q -1 )\log
         u_i$. So $S_1 = - \sum_i u_i \log u_i $, which is Shannon
         entropy.
      \end{definition}
      Tsallis entropy  \cite{Tsallis1988Entropy}
      can also be viewed as a $ q$-deformation of Shannon entropy
      because $ S_q (\bu) = - \sum_i u_i \log_q u_i $  
      where $ \log_q (u) = \frac{ u^{1-q } -1 }{ 1 - q } $
      is the $ q$-logarithm. Clearly 
      $  \log_q (u) \rightarrow \log (u)$ 
      when $q \rightarrow 1 $.

      In summary, we conclude that although the primal problems of
      boosting with different loss functions seem dissimilar, their
      corresponding dual problems share the same formulation. 
      Most of them can be
      interpreted as entropy regularized \lpboost.
      Table~\ref{TAB:entr} 
      summarizes the result. 
      The analysis of \logitboost will be presented in the next
      section. 

%
%
%
%
         \begin{table*}[th!]
         \caption
         {
               Dual problems of boosting algorithms are entropy
               regularized \lpboost. 
         }
         \centering
         \begin{small}
         \begin{tabular}{l l l}
         \hline\hline
algorithm  &  loss in primal          & entropy regularized \lpboost
in dual  \\
\hline 
\adaboost   & exponential loss & Shannon entropy \\
\logitboost & logistic loss    & binary relative entropy  \\
soft-margin $ \ell_p (p>1)$ \lpboost & generalized hinge loss & Tsallis entropy \\ 
         \hline\hline
         \end{tabular}
         \end{small}
         \label{TAB:entr}
         \end{table*}


\subsection{Lagrange dual of \logitboost}
\label{sec:logitboost}

      Thus far, we have discussed \adaboost and its relation to
      \lpboost.  In this section, we consider \logitboost
      \cite{Friedman2000Additive} from its dual.

      \begin{theorem}
          The dual of \logitboost is a binary relative entropy
          maximization problem,
          which writes
         \begin{align}
         \label{EQ:dual_logit0}
         \smaximize_{r, \bu} ~ &
         \frac{r}{T} -  {\textstyle \sum_{i=1}^M} \left[
                { (- u_i ) \log (-u_i) } + ( 1 +u_i)
         \log( 1+u_i) 
                                         \right]
         \notag \\
         \sst ~ & {\textstyle \sum_{i=1}^M} { u_i y_i H_{ij} }
         \geq  r 
         ~(\forany j=1,\cdots,N).
         \end{align}
      \end{theorem}
      We can also rewrite it into an equivalent form:
         \begin{align}
         \label{EQ:dual_logit1}
         \sminimize_{r, \bu} ~ &
          r +  T {\textstyle \sum_{i=1}^M} \left[
                {  u_i  \log u_i } + ( 1 - u_i)
         \log( 1 - u_i) 
                                         \right]
         \notag \\
         \sst ~ & {\textstyle \sum_{i=1}^M} { u_i y_i H_{ij} }
         \leq  r 
         ~(\forany j=1,\cdots,N).
         \end{align}
      The proof follows the fact that the conjugate of the logistic
      loss function $\logit( x ) = \log ( {1 + \exp - x } ) $ is
      \[
            \logit^*( u ) = 
              \begin{cases}
                     (- u ) \log (- u ) + ( 1 + u ) \log ( 1 + u), 
                                 & \!\!\!\!\! 0 \geq u \geq -1;\\
                              \infty,
                                 & \!\!\!\!\! \text{otherwise}, 
              \end{cases}
      \]
      with $ 0 \log 0 = 0 $. 
      $ \logit^*(u) $ is a convex function in its domain.
      The corresponding primal is 
         \begin{align}
         \label{EQ:primal_logit1}
         \min_{\bw} ~ &
         {\textstyle \sum_{i=1}^M} 
                \logit( z_i)
         \notag \\
         \sst ~ & z_i =  y_i H_i \bw,
         ~(\forany i=1,\cdots,M),
          \\
          ~ & \bw \psd \b0,  \b{1}^\T \bw = \tfrac{1}{T}. \notag
         \end{align}
      %
      %
      %
      %
      In \eqref{EQ:dual_logit1}, the dual variable $ \bu $ has a
      constraint
      $  \b{1} \psd \bu \psd \b{0} $, which is automatically enforced by the
      logarithmic function. 
       Another difference of \eqref{EQ:dual_logit1} 
       from duals of \adaboost and \lpboost \etc is that $ \bu $ does
       not need to be normalized. In other words, in \logitboost the
       weight associated  with each training sample is not necessarily
       a distribution.
      As in \eqref{EQ:weight2} for \adaboost,  
      we can also relate a dual optimal point $ \bu^\star $ and a
      primal optimal point $ \bw^\star $ ( between
      \eqref{EQ:dual_logit1} and
      \eqref{EQ:primal_logit1} ) by
      \begin{equation}
         u_i^\star =    \frac{ \exp - z_i^\star } 
                          { 1 + \exp - z_i^\star }, \,\, \forany i = 1,\cdots
                          M.
         \label{EQ:logit_w1}
      \end{equation}
       So the margin of $ \bx_i $ is solely determined by $ u_i^\star$:
       $  z_i^\star =  \log {\frac{ 1 - u_i^\star }{ u_i^\star } } $,
       $ \forany i$. For a positive margin ($ \bx_i$ is 
       correctly classified), we must have $ u_i^\star < 0.5 $.

       Similarly, we can also use CG to solve \logitboost.  As shown
       in Algorithm~\ref{alg:AdaCG} in the case of AdaBoost,  the only
       modification is to solve a different dual problem (here we need
       to solve \eqref{EQ:dual_logit1}).

\subsection{AdaBoost$_{\ell1}$ approximately maximizes the average margin and
minimizes the margin variance}

Before we present our main result, a lemma is needed. 
            
\begin{lemma}
      The margin of AdaBoost$_{\ell1}$ and AdaBoost follows the Gaussian distribution.
      In general, the larger the number of weak classifiers, the more
      closely does the margin follow the form of Gaussian under the
      assumption that selected weak classifiers are uncorrelated.  
\label{Lem:1}
\end{lemma}
\begin{proof}
   The central limit theorem \cite{Kallenberg1997Foundations} states
   that the sum of a set of i.i.d.~\-r\-a\-n\-dom variables $ x_i
   $, ($i=1\cdots N $) is
   approximately distributed following a Gaussian distribution if the
   random variables have finite mean and variance.

   \begin{figure}[t!]
                \begin{center}
                  \includegraphics[width=0.35\textwidth]{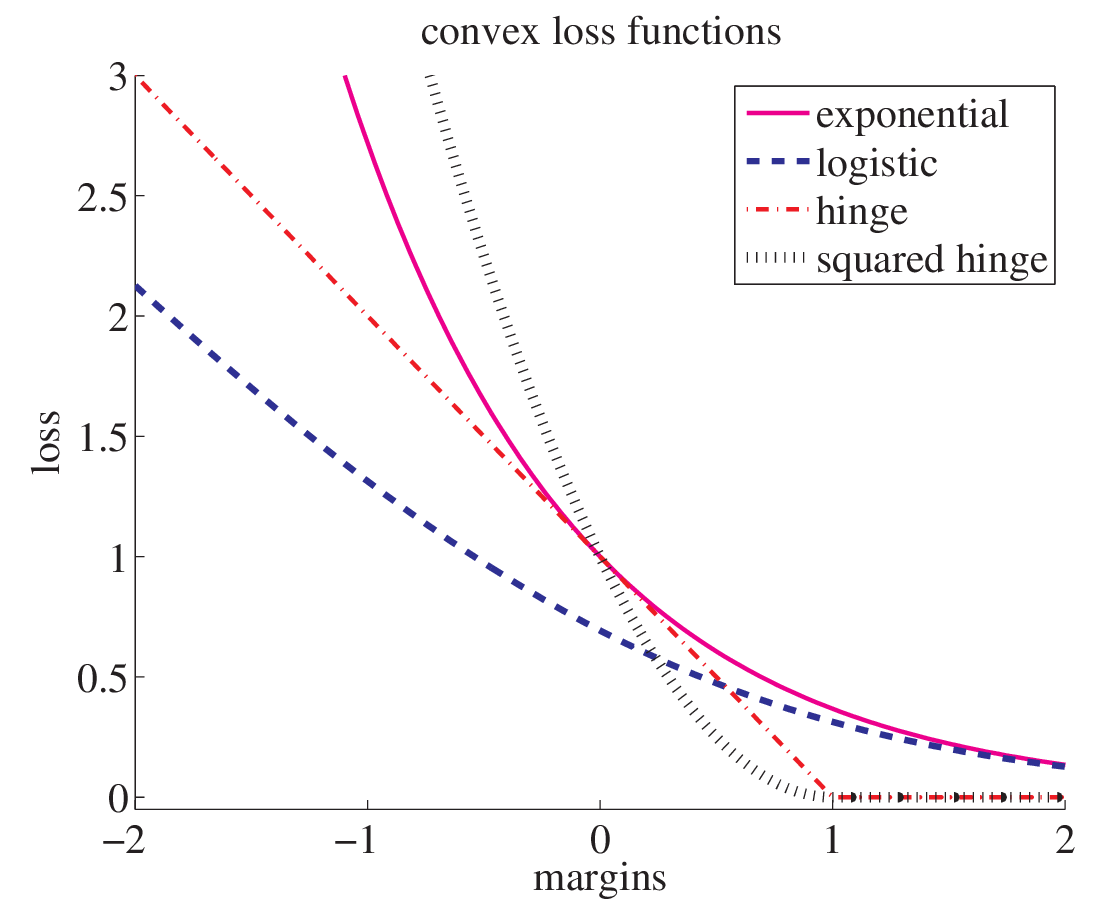}
                \end{center}
                \caption{Various loss functions used in
                classification.
                Exponential: $ \exp -s  $;
                logistic: $ \log ( 1 + \exp -s ) $;
                hinge: $\smaximize \{0, 1 - s \} $;
                squared hinge: $(\smaximize \{0, 1 - s \})^2$.
                Here $s = y F(\bx)$.
                }
                \label{fig:lossfun}
   \end{figure}

   Note that the central limit theorem applies when each variable
   $x_i$ has an {\em arbitrary}  probability distribution $ \cQ_i$
   as long as the mean and variance of $ \cQ_i $ are finite.    
      
   As mentioned, the normalized margin of \adaboost for $i$-th example
   is defined as 
   \begin{equation}  
         \margin_i = ( y_i 
         {
         \textstyle
         \sum_{j=1}^N h_j (\bx_i) w_j 
         } )
            / \b{1}^\T \bw
         = - z_i / \b{1}^\T \bw.
   \label{EQ:margin}
   \end{equation}
   In the following analysis, we ignore the normalization term $
   \b{1}^\T \bw $ because it does not
   have any impact on the margin's distribution. 
   Hence the margin $ \varrho $ is the sum of $ N $ variables 
   $ \hat w_j $ with $ \hat w_j = y_i 
      h_j (\bx_i) w_j $. 
   It is easy to see that each $ \hat w_j $ follows a discrete
   distribution with binary values either 
   $ w_j $ or $ -w_j $. Therefore $w_j$
   must have  finite mean and variance.
   Using the central limit theorem, we know that 
   the distribution of $ \varrho_i $ is a Gaussian.

    In the case of discrete variables ($ \varrho_i $ can be discrete), 
    the assumption
    identical distributions can be substantially weakened \cite{Feller1968Prob}. 
    The generalized central limit theorem 
    essentially states that anything that can be thought of as being
    made up as the sum of many small independent variables
    is approximately normally
    distributed.

   A condition of the central limit theorem is that the $ N $
   variables must be independent. 
   In the case of the number of weak hypotheses is finite, 
   as the margin is expressed in \eqref{EQ:margin},
   each $ h_j ( \cdot )$ is fixed beforehand, and assume all the training
   examples are randomly independently drawn, the variable $ \varrho_i$ 
   would be independent too.
   When the set of weak hypotheses is infinite, 
   it is well known
   that usually \adaboost selects independent
   weak classifiers such that each weak classifier makes different
   errors on the training dataset \cite{Meir2003Boosting}. 
   In this sense,
   $ w_j $ might be viewed as roughly independent from each
   other. 
   More diverse weak classifiers will make the selected weak
   classifiers less dependent.\footnote{Nevertheless, 
   this statement is not
   rigid.} 
\end{proof}

\begin{figure}[t]
   \begin{center}
      \includegraphics[width=0.35\textwidth]{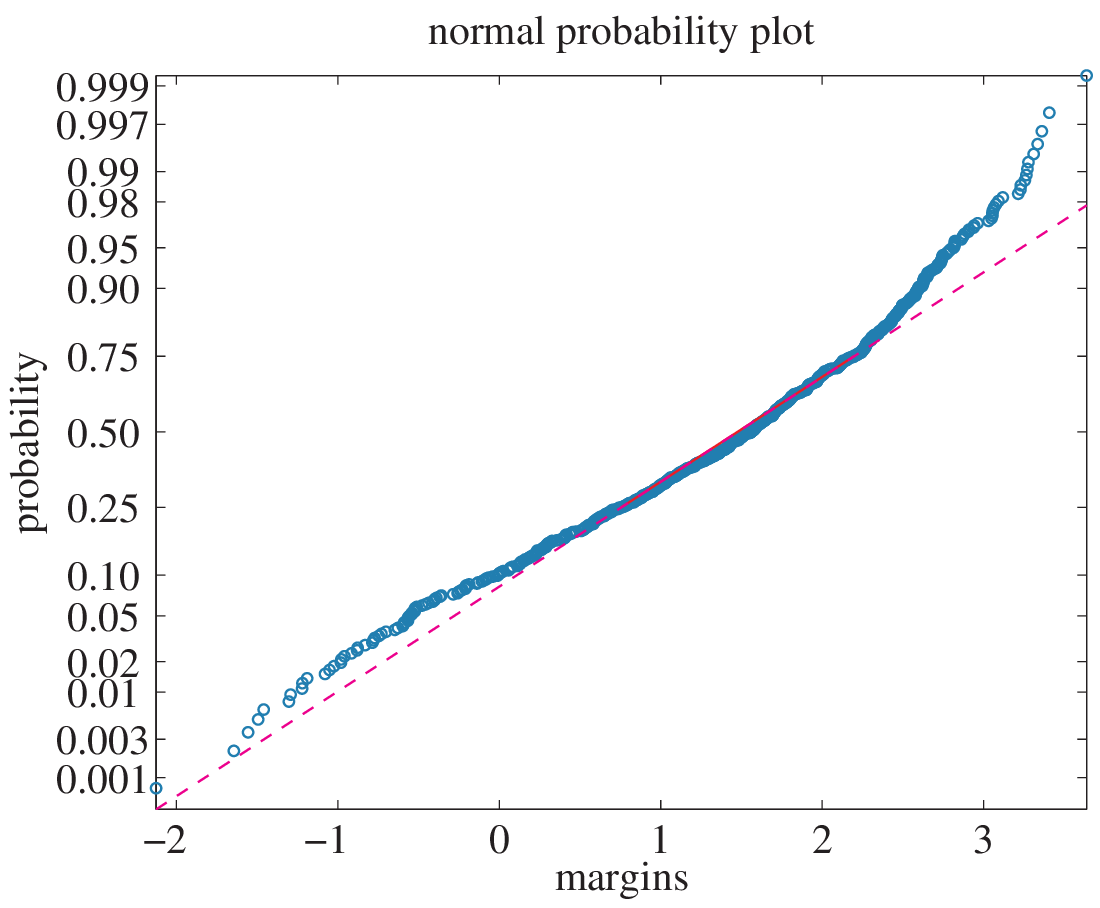}
      \includegraphics[width=0.35\textwidth]{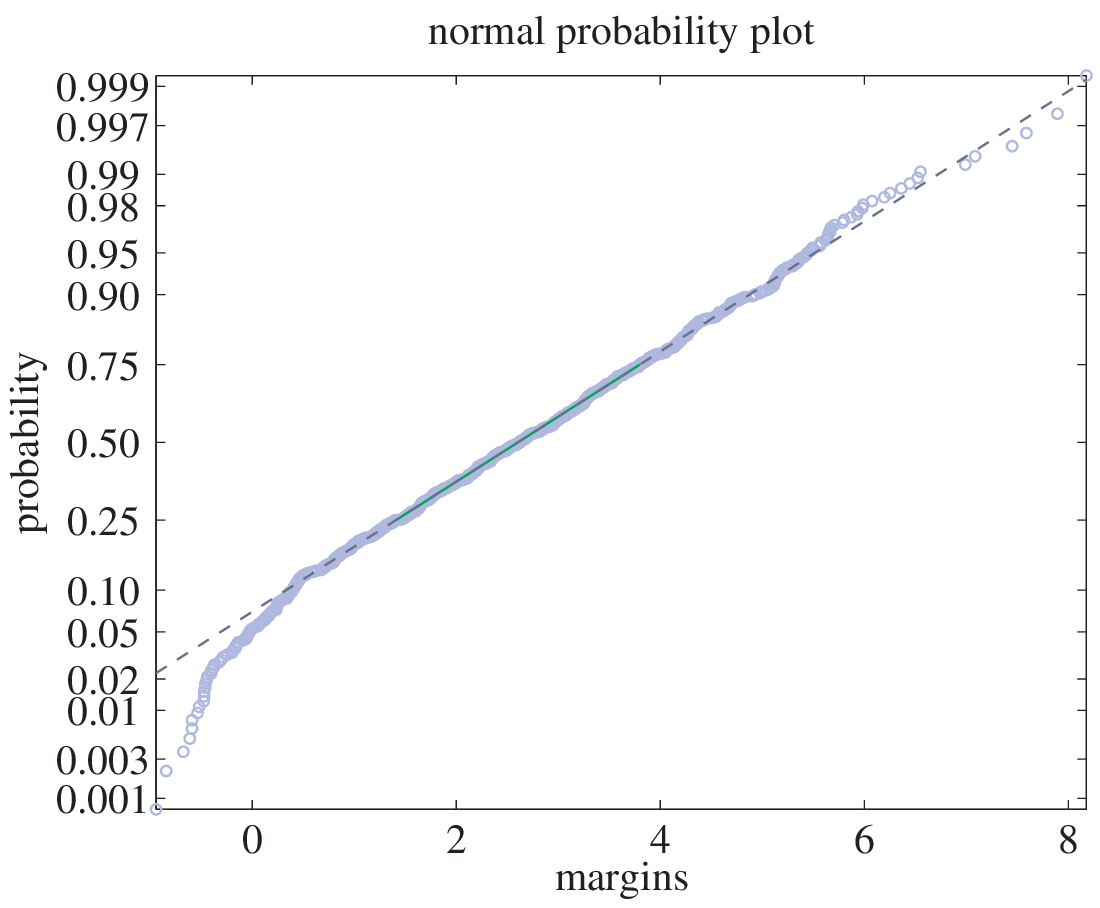}
   \end{center}
   \caption{Gaussianity test for the margin distribution with
    $ 50 $ and $ 1100 $ weak classifiers, respectively.
    A Gaussian distribution will form a straight line.
    The dataset used is {\em australian}. 
    } 
   \label{fig:margin_gauss}
\end{figure}

      Here we give some empirical evidence for approximate
      Gaussianity.  The normal (Gaussian) probability plot is used to
      visually assess whether the data follow a Gaussian distribution.
      If the data are Gaussian the plot forms a straight line.
      Other distribution
      types introduce curvature in the plot.
      We run \adaboost with decision stumps on the dataset {\em
      australian}.
      \fig~\ref{fig:margin_gauss} shows two plots of the margins with
      $ 50 $ and $ 1100 $ weak classifiers, respectively. 
      We see that with $ 50 $ weak classifiers, the margin
      distribution can be reasonably approximated by a Gaussian;
      with $ 1100 $ classifiers, the distribution is very close to a
      Gaussian.  
      The kurtosis of a 1D data provides a
      numerical evaluation of the Gaussianity. We know that the
      kurtosis of a Gaussian distribution is zero and almost all the
      other distributions have non-zero kurtosis. In our experiment,
      the kurtosis is $ -0.056 $ for the case with $ 50 $ weak
      classifiers and $ -0.34 $ for $ 1100 $ classifiers. Both are
      close to zero, which indicates \adaboost's margin distribution
      can be well approximated by Gaussian.

\begin{theorem}
   \label{thm:4}
      AdaBoost$_{\ell1}$ approximately maximizes the unnormalized
      average margin and at the same time
      minimizes the variance of the margin distribution under the
      assumption that the margin follows a Gaussian distribution.
\end{theorem}

\begin{proof}
   From \eqref{EQ:2A} and \eqref{EQ:margin},
   the cost function that AdaBoost$_{\ell1} $ minimizes is
   \begin{equation}
   \Fada( \bw ) = 
                     \log \left(  
                     {\textstyle 
                     \sum_{i=1}^M } \exp -  \frac{ \varrho_i }{ T }
                     \right). 
   \label{EQ:Ada3}
   \end{equation} 
   %
   %
   %
   %
   As proved in Lemma~\ref{Lem:1}, $ \varrho_i $ follows a Gaussian
   \[ 
         \cG ( \varrho; \varrhomean, \sigma ) = 
         {1 \over \sqrt{2 \pi} \sigma }
               \exp - \frac{ ( \varrho - \varrhomean )^2 }{ 2 \sigma^2 }, 
   \]
   with mean $ \varrhomean $, variance $ \sigma^2 $; and $ \sum_{i=1}^M
   \varrho_i = 1$. We assume that the optimal value of the regularization
   parameter $ T $ is known a priori. 
   
   The Monte Carlo integration method can be used to compute a continuous
   integral 
   \begin{equation}
     \int g( x ) f(x) \dif x 
     \simeq  \frac{1}{K} \sum_{k=1}^K f(x_k), 
   \end{equation}
   where $ g(x) $ is a probability distribution such that $ \int g(x)
   \dif x =1 $ and $ f(x) $ is an arbitrary function.
   $ x_k$, $(k=1\cdots K)$, are randomly sampled from the distribution 
   $ g(x) $. The more samples are used, the more accurate the
   approximation is.     
   
   \eqref{EQ:Ada3}
   can be viewed as a discrete Monte Carlo approximation
   of the following integral (we omit a constant term $ \log M $,
   which is irrelevant to the analysis):
   \begin{align}
& \hat{\Fada}(\bw)  \notag
       \\
       &= \log 
          \int_{ \varrho_1 }^{ \varrho_2 } \cG (  \varrho; \varrhomean, \sigma )
          \exp \left( -\frac{ \varrho } { T } \right)
           \dif \varrho
          \notag \\
       &= \log 
          \int_{ \varrho_1 }^{ \varrho_2 } 
         {1 \over \sqrt{2 \pi} \sigma }
                     \exp
                     \left( 
                     - 
                    \frac{ ( \varrho - \varrhomean )^2 }{ 2 \sigma^2 }
                    -  \frac{\varrho}{ T }
                    \right)
           \dif \varrho
           \notag \\
      &= \log
      \left[
      \tfrac{1}{2} \exp \left(
           - \frac{ \varrhomean }{ T } +  \frac{\sigma^2}{2 T^2}  
           \right)
      \erf \left(
             \frac{\varrho - \varrhomean}{\sqrt{2} \sigma } +
      \frac{\sigma}{ \sqrt{2} T}   
      \right)
       \bigg|_{\varrho_1}^{\varrho_2}
              \bigg.
      \right]
              \notag \\
      &=
            - \log 2 - \frac{ \varrhomean }{ T } +  \frac{\sigma^2}{2 T^2}
            + \log \left[
              \erf \left(
                     \frac{\varrho - \varrhomean}{\sqrt{2} \sigma } +
                     \frac{\sigma}{ \sqrt{2} T}
                  \right)\bigg|_{\varrho_1}^{\varrho_2}
              \bigg. 
              \right],
      \label{EQ:Ada4}
   \end{align}
         where $ 
         \erf( x ) = \frac{2}{\sqrt{ \pi } } \int_0^x \exp - s^2 
         \dif s 
         $
         is the Gauss error function.
         The integral
         range is $ [ \varrho_1, \varrho_2 ]$. 
         With no explicit knowledge about the integration range, we
         may roughly calculate the integral from $ -\infty$ to $
         +\infty $. 
         Then the last term in \eqref{EQ:Ada4} is $ \log 2$
         and the result is analytical and simple
         \begin{equation}
             \hat{\Fada}(\bw) = - \frac{\varrhomean}{T } + 
                          \frac{1}{2} \frac{\sigma^2}{T^2}.
            \label{EQ:Ada5}
         \end{equation}
         This is a reasonable approximation because 
         Gaussian distributions drop off quickly (Gaussian is not
         considered a heavy-tailed distribution). 
         Also this approximation implies that
         we are considering the case that the number of samples goes to $ + \infty $.

         Consequently, \adaboost approximately 
         maximizes the cost function 
         \begin{equation}
             - \hat{\Fada}( \bw ) =    \frac{\varrhomean}{T } - 
                           \frac{1}{2} \frac{\sigma^2}{T^2}.
            \label{EQ:Ada7}
         \end{equation}
         This cost function has a clear and simple interpretation: The first
         term $ {\varrhomean} / {T} $ is the unnormalized average margin
         and the second term $ {\sigma^2} / {T^2} $
         is the unnormalized margin variance.
         So \adaboost maximizes the unnormalized average margin and
         also takes minimizing the unnormalized margin variance
         into account.
         This way a better {\em margin distribution}
         can be obtained. 
\end{proof}

      Note that Theorem~\ref{thm:4} depends on 
      Lemma \ref{Lem:1} that does not necessarily hold in practice. 

      Theorem~\ref{thm:4} is an important result in the sense that 
      it tries to 
      contribute to the open question why \adaboost works so well. 
      %
      %
      %
      Much previous work intends to believe that \adaboost maximizes the
      minimum margin.
      %
      %
      %
      We have theoretically shown that AdaBoost$_{\ell1}$
      optimizes the entire
      margin distribution by maximizing the mean and minimizing the
      variance of the margin distribution.

      We notice that when $ T \rightarrow 0 $,
      Theorem~\ref{thm:4} becomes invalid because 
      the Monte Carlo integration cannot approximate
      the cost function of \adaboost \eqref{EQ:Ada3} well.
      In practice, $ T $ cannot approach to zero arbitrarily in
      \adaboost.

      One may suspect that Theorem~\ref{thm:4} contradicts the
      observation of similarity between \lpboost and \adaboost as
      shown in Section~\ref{sec:adalp}. \lpboost maximizes the minimum
      margin and the dual of \adaboost is merely an entropy
      regularized \lpboost. At first glance, the dual variable $ r $
      in \eqref{EQ:dual_ada1}, \eqref{EQ:dual_LP1}, and
      \eqref{EQ:dual_SoftLP1} should have the same meaning, \ie,
      maximum edge, which in turn corresponds to the minimum margin in
      the primal. Why average margin?
      To answer this question, let us again take a look at
      the optimality conditions. 
      Let us denote the optimal values of \eqref{EQ:dual_ada1} 
      $  r^\star  $ and $ \bu^\star $.
      At convergence, we have 
      $ 
      \frac{1}{T} ( - r^\star + T \sum_{i=1}^M u_i^\star \log  u_i^\star
                    )
                    =  \opt{\ref{EQ:dual_ada0}}
                        =\opt{\ref{EQ:2}} 
                        =\opt{\ref{EQ:2A}}.
      $
      Hence, we have
      \[
       r^{\star} = T \sum_{i=1}^M u^{\star} \log  u^{\star}
      - T \log \left( \sum_{i=1}^M \exp - \frac{\margin_i^\star}{T}
      \right),
      \]
      where $ \margin_i^\star$ is the normalized margin for $ \bx_i $.
      Clearly this is very different from the optimality conditions of
      \lpboost, which shows that $ r^\star$ is the minimum margin. 
      Only when $ T \rightarrow 0 $, the above relationship 
      reduces to $ r^\star =  \sminimize_i \{ \margin_i^\star \}
      $---same as the case of \lpboost. 
      %
      %
      %


   \subsection{\adaboost-QP: Direct optimization of the margin mean
   and variance using quadratic programming}

   \label{sec:QP}

   The above analysis suggests that we can directly optimize the
   cost function \eqref{EQ:Ada7}. In this section we show that
   \eqref{EQ:Ada7} is a convex programming (more precisely, quadratic
   programming, QP) problem in the variable $ \bw $ if we know all the
   base classifiers and hence it can be efficiently solved.
   Next we formulate the QP problem in detail. We call the proposed
   algorithm \adaboost-QP.\footnote{In \cite{Ratsch2001Soft}, 
   the authors proposed QP$_{\rm reg}$-\adaboost for soft-margin \adaboost
   learning, which is inspired by SVMs.
   Their QP$_{\rm reg}$-\adaboost is completely different from ours. 
   }

    In kernel methods like SVMs, the original space $
    \cX$ is mapped to a feature space $ \cF $. The mapping function $
    \Phi(\cdot) $ is not explicitly computable. 
%
%
%
   It is shown in \cite{Ratsch2002BoostSVM} that in boosting, one can
   think of the mapping function $ \Phi(\cdot)$ being {\em explicitly}
   known:
   \begin{equation}
      \Phi(\bx): \bx \mapsto [ h_1(\bx),\cdots,h_N(\bx) ]^\T,
      \label{EQ:mappingfun}
   \end{equation}
   using the weak classifiers. Therefore, any 
   weak classifier set $ \cH $ spans a feature space $ \cF $. 
   We can design an algorithm that optimizes \eqref{EQ:Ada7}:
      \begin{equation}
         \sminimize_\bw  \,\, \tfrac{1}{2} \bw^\T  A  \bw - 
         T
         \bb^\T \bw,
         \,\,\,\,
         \sst\,\, \bw \psd 0, \b{1}^\T \bw = 1,
      \end{equation}
      where
      $\bb = \frac{1}{M}\sum_{i=1}^M y_i H_i^\T
      = \frac{1}{M}\sum_{i=1}^M y_i \Phi(\bx_i) 
      $,
      and 
      $
      A = \frac{1}{M}\sum_{i=1}^M (  y_i H_i^\T - \bb )
            (  y_i H_i^\T - \bb )^\T
        = \frac{1}{M}\sum_{i=1}^M (  y_i \Phi(\bx_i) - \bb )
        (  y_i \Phi(\bx_i) - \bb )^\T.
      $\footnote{To show the connection of \adaboost-QP with kernel methods,
      we have written $ \Phi(\bx_i) = H_i^\T$.}
      Clearly $ A $ must be positive semidefinite and  this is a
      standard convex QP problem. The non-negativeness constraint $ \bw
      \psd 0 $ introduces sparsity as in SVMs. Without this
      constraint, the above QP can be analytically solved using
      eigenvalue decomposition---the largest eigenvector is the
      solution.
      Usually all entries of this solution would be active (non-zero
      values). 
      
      In the kernel space, 
      \[
      \bb^\T\bw = \tfrac{1}{M} \left( \sum\nolimits_{y_i =1} 
      \Phi(\bx_i) - \sum\nolimits_{y_i = -1} \Phi(\bx_i) \right)^\T
      \bw
      \]
      can be viewed as the projected $ \ell_1$ norm distance between
      two classes because typically this value is positive assuming
      that each class has the same number of examples. 
      The matrix $ A $ {\em approximately} 
      plays a role as the total scatter
      matrix in kernel linear discriminant analysis (LDA). 
      Note that  \adaboost does not take the number of examples in
      each class into consideration when it models the problem. 
      In contrast, LDA (kernel LDA) takes training example number into
      consideration.
      This may explain why an LDA post-processing on \adaboost
      gives a better
      classification performance on face detection
      \cite{Wu2005Linear}, which is a highly
      imbalanced classification problem. 
      This observation of similarity between \adaboost and kernel LDA
      may inspire new algorithms. 
%
%
%
%
      We are also interested in developing a CG based
      algorithm for iteratively generating weak classifiers.

%
%

\subsection{\adaboost-CG: Totally corrective \adaboost using column generation}
\label{sec:adaboost-cg}

      The number of possible weak classifiers may be infinitely large.
      In this case it may be infeasible to solve the optimization {\em
      exactly}. 
      \adaboost works on the primal problem directly by switching
      between the estimating weak classifiers and computing optimal
      weights in a coordinate descent way.  
      There is another method for working out of this
      problem by using an optimization technique termed column
      generation (CG) \cite{Lubbecke2005Selected,Demiriz2002LPBoost}.
      CG mainly works on the dual problem.   
      The basic concept of the CG method is to add
      one constraint at a time to the dual problem until an optimal
      solution is identified.  More columns
      need to be generated and added to the problem to achieve optimality.
      In the primal space, the
      CG method solves the problem on a subset of variables, which
      corresponds to a subset of constraints in the dual. 
      When a column is not included in the primal, the corresponding
      constraint does not appear in the dual. 
      That is to say, a relaxed version of the dual problem is solved. 
      If a constraint absent
      from the dual problem is violated by the solution to the
      restricted problem, this constraint  needs to
      be included in the dual problem to further restrict its feasible
      region.  
      In our case,
      instead of solving the optimization of
      \adaboost directly, one computes the most violated constraint in
      \eqref{EQ:dual_ada1} iteratively for the current solution and
      adds this constraint to the optimization problem.  In theory,
      any column that violates dual feasibility can be added.  To do
      so, we need to solve the following subproblem:
      \begin{equation}
        h' ( \cdot ) =  \argmax_{h( \cdot ) } ~ 
                \textstyle  \sum_{i=1}^M u_i y_i h ( \bx_i).
         \label{EQ:pickweak}
      \end{equation}
   This strategy is exactly the same as the one that stage-wise
   \adaboost and
   \lpboost use for generating the best weak classifier. That is, to
   find the weak classifier that produces minimum weighted training
   error.  Putting all the above analysis together, we summarize our
   AdaBoost-CG in Algorithm~\ref{alg:AdaCG}.

   The CG optimization (Algorithm~\ref{alg:AdaCG}) is
   so general  that it can be applied to all the boosting algorithms
   consider in this paper by solving the corresponding dual.   
   The convergence follows general CG algorithms, which is easy to
   establish.
   When a new $ h'(\cdot) $ that violates dual feasibility is added,
   the new optimal value of the dual problem (maximization) would
   decrease.  Accordingly, the optimal value of its primal problem
   decreases too because they have the same optimal value due to zero
   duality gap. Moreover the primal cost function is convex, therefore
   eventually it converges to the global minimum. 
   A comment on the last step of Algorithm~\ref{alg:AdaCG} is that we
   can get the value of $ \bw $ easily. 
         Primal-dual interior-point (PD-IP) methods work on the primal
         and dual problems simultaneously and therefore both primal
         and dual variables are available after convergence. We use
         \mosek \cite{Mosek}, which implements PD-IP methods. The
         primal variable  $ \bw $ is obtained {\em for free} when
         solving the dual problem  \eqref{EQ:dual_ada1}.

   The dual subproblem we need to solve has one constraint added at
   each iteration. Hence after many iterations solving the dual
   problem could become intractable in theory. In practice,
   \adaboost-CG converges quickly on our tested datasets. As pointed
   out in \cite{Sonnenburg2006Large}, usually only a small number of
   the added constraints are active and those inactive ones may be
   removed. This strategy prevents the dual problem from growing too
   large.

   \adaboost-CG is totally-corrective in the sense that the
   coefficients of all weak classifiers are updated at each iteration.
   In \cite{Sochman2004AdaBoost}, an additional correction procedure
   is inserted to \adaboost's weak classifier selection cycle for 
   achieving totally-correction. The inserted correction procedure 
   aggressively reduces the {\em upper bound} of the training error. 
   Like \adaboost, it works in the primal. In contrast, our
   algorithm optimizes the regularized loss function directly and
   mainly works in the dual space. 
   In \cite{Warmuth2006Total}, a totally-corrective boosting is proposed by
   optimizing the entropy, which is inspired by
   \cite{Kivinen1999Boosting}. As discussed, no explicit  
   primal-dual connection is established. That is why an LPBoost procedure
   is needed over the obtained weak classifiers 
   in order to calculate the primal variable $ \bw $.  
   In this sense, \cite{Warmuth2006Total} is also similar to the work of 
   \cite{Ratsch2001Soft}.

   \linesnumbered
   \begin{algorithm*}[t]
     \caption{\adaboost-CG.}   
   \begin{algorithmic}  
   \KwIn{Training set  $(\bx_i, y_i), i = 1\cdots M$;
         termination threshold $ \varepsilon > 0$; regularization
         parameter $ T $; (optional) maximum iteration
         $N_\mathrm{max}$.
   }
        { {\bf Initialization}:
        \begin{enumerate}
        \item
           $ N = 0 $ (no weak classifiers selected);
        \item
           $ \bw = {\bf 0} $ (all primal
               coefficients are zeros); 
        \item 
            $ u_i = \frac{1}{ M }$, 
            $ i = 1
            \cdots M$ (uniform dual weights). 
        \end{enumerate}
   }

   \While{ ${\mathrm{true}}$ }
   {
      \begin{enumerate}
      \item
         Find a new base $ h'(\cdot) $ by solving Problem
         \eqref{EQ:pickweak};
     \item
         Check for optimal solution: \\
         {\bf if}{ $
                    \sum_{ i=1 }^M u_i y_i h' ( \bx_i )  
                           < r + \varepsilon $},
                           { \bf then}
           break (problem solved); 
     \item
         Add $  h'(\cdot) $ to the restricted master problem, which
         corresponds to a new constraint in the dual;
     %
     %
     %
     \item
         Solve the dual to obtain updated $ r $ and
         $ u_i$ ($ i = 1,\cdots,M$):
               for \adaboost, the dual is \eqref{EQ:dual_ada1};
     \item
         $N = N + 1$ (weak classifier count);
     \item
         (optional) 
         {\bf if}{ $ N \geq N_\mathrm{max} $}, {\bf then} break (maximum
         iteration reached).    
     \end{enumerate}
   }
   \KwOut{
      \begin{enumerate}
      \item
         Calculate the primal variable $ \bw $ from the optimality
         conditions and the last solved dual problem;
      \item
         The learned classifier 
         $ F ( \bx ) = \sum_{j=1}^{N} w_j h_j( \bx ) $.
      \end{enumerate}
   }
   \end{algorithmic}
   \label{alg:AdaCG}
   \end{algorithm*}

The following diagram summarizes the relationships that we have
derived on the boosting algorithms that we have considered.
\begin{equation*}
   \begin{CD}
       \text{AdaBoost$_{\ell1}$ primal}  @>\text{\adaboost-CG}>\text{Lagrange
       duality}>\text{AdaBoost$_{\ell1}$ dual}   \\
@V\text{Theorem}~\ref{thm:4}VV     @A\text{entropy}A{\text{ regularization}}A   \\
      \text{\adaboost-QP}      @.       \text{\lpboost dual}  \\
      \end{CD}
\end{equation*}


\section{Experiments}
\label{sec:exp}

      In this section we provide experimental results to verify the
      presented theory. 
      We have mainly used decision stumps as weak classifiers due to
      its simplicity and well-controlled complexity. 
      In some cases, we have also used one of the simplest linear
      classifiers, LDA, as weak classifiers. To avoid the singularity
      problem when solving LDA, we add a scaled identity matrix $
      10^{-4} \I $ to the within-class matrix. 
      For the CG optimization framework, we have confined ourself to
      \adaboost-CG although the technique is general and applicable
      for optimizing other boosting algorithms.

\subsection{\adaboost-QP}
\label{sec:qp_exp}

      We compare \adaboost-QP against  \adaboost.
      We have used $ 14 $ benchmark datasets \cite{LIBSVMdata2001}. 
Except {\em mushrooms}, 
       {\em svmguide1},
       {\em svmguide3} and {\em w1a},
       all the other datasets have been scaled to 
       $[-1, 1]$.
      We randomly split each dataset into training, cross-validation
      and test sets at a ratio of $70:15:15$.

      The stopping criterion of \adaboost is determined by
      cross-validation on $\{ 600, 800, 1000, 1200, 1500\}$ rounds of
      boosting. 
      For \adaboost-QP, the best value for the parameter $ T $ is chosen from 
      $\{
      \frac{1}{10}, 
      \frac{1}{20}, 
      \frac{1}{30}, 
      \frac{1}{40},
      \frac{1}{50}, 
      \frac{1}{100}, 
      \frac{1}{200}, 
      \frac{1}{500} 
      \}$ by cross-validation.  In this experiment, decision stumps
      are used as the weak classifier such that the complexity of the
      base classifiers is well controlled.

      \adaboost-QP must access all weak classifiers {\em a priori}.
      Here we run \adaboost-QP on the $1500$ weak classifiers
      generated by \adaboost. Clearly this number of hypotheses may
      not be optimal. Theoretically the larger the size of the weak
      classifier pool is, the better results \adaboost-QP may
      produce.  
      Table~\ref{TAB:QP1} reports the  results. 
      The experiments show
      that among these $ 14 $ datasets, \adaboost-QP outperforms
      \adaboost on $ 9 $ datasets in terms of generalization error.
      On {\em mushrooms}, both perform
      very well. On the other $ 4 $ datasets, \adaboost is better. 
      
%
       We have also computed the normalized version of the cost
       function value of \eqref{EQ:Ada7}. 
       In most cases \adaboost-QP has a larger value. 
       This is not
       surprising since \adaboost-QP directly maximizes
       \eqref{EQ:Ada7} while \adaboost approximately maximizes it.
       Furthermore, the normalized loss function value is close to
       the normalized average margin because the
       margin variances for most datasets are very small compared with
       their means.

\begin{figure}[t!]
                \begin{center}
                  \includegraphics[width=0.35\textwidth]{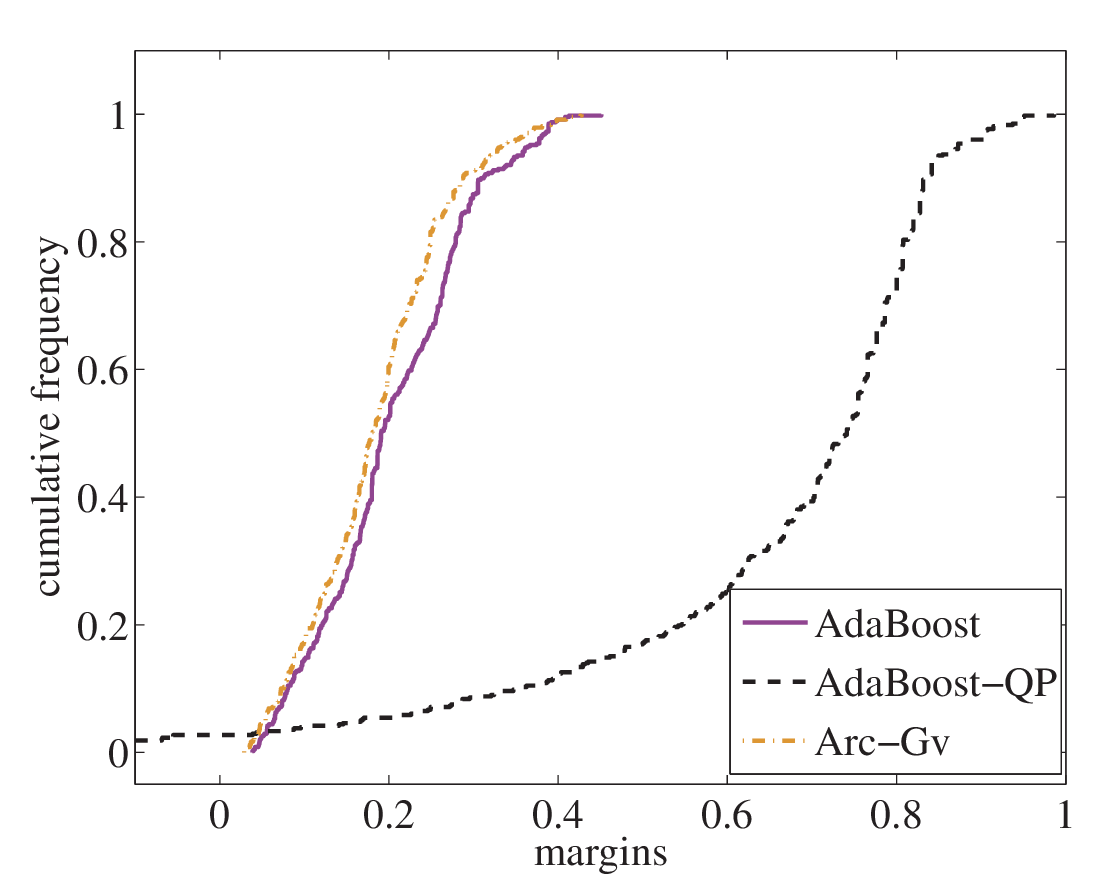}
                \end{center}
                \caption{Cumulative 
                margins for \adaboost, \adaboost-QP and
                \arcgv for the
                breast cancer dataset
                using  decision stumps. 
                Overall, the margin distribution of 
                \adaboost-QP is the best and it has a smallest test
                error. 
                \adaboost and \arcgv run
                $600$ rounds of boosting. Test error for
                \adaboost, \adaboost-QP and \arcgv is
                $ 0.029,  0.027, 0.058$ respectively. 
                }
                \label{fig:QP1}
             \end{figure}

      We also compute the largest minimum margin and average margin on
      each dataset. 
      On all the datasets \adaboost has
      a larger minimum margin than \adaboost-QP,
      This confirms that
      the minimum margin is not crucial for the generalization error.
      On the other hand,
      the average margin produced by \adaboost-QP,
      which is the first term of the cost function
      \eqref{EQ:Ada7}, is consistently larger than the one obtained by
      \adaboost. 
      Indirectly, we have shown that a better overall margin
      distribution is more important than the largest minimum margin.
      In \fig~\ref{fig:QP1} we plot 
      cumulative margins for \adaboost-QP and \adaboost on
      the breast-cancer dataset with decision stumps.
      We can see that while \arcgv has a largest minimum
      margin, it has a worst margins distribution overall.
      If we examine the average margins, \adaboost-QP is
      the largest; \adaboost seconds and \arcgv is least. 
      Clearly a better overall distribution does lead to  a smaller
      generalization error.
      When \arcgv and  \adaboost run for more rounds, their margin
      distributions seem to converge.  
      That is what we see in \fig~\ref{fig:QP1}.
      These results agree well with our theoretical analysis 
      (Theorem~\ref{thm:4}). 
      Another observation from this experiment
      is that, to achieve the same performance,
      \adaboost-QP tends to use fewer weak classifiers than
      \adaboost does.

      We have also tested \adaboost-QP on full sets of weak
      classifiers because the number of possible decision stumps is
      finite (less than (number of features $- 1$)  $\times$ 
      (number of examples)). 
      Table~\ref{TAB:QP2} reports the test error of \adaboost-QP 
      on some small datasets. As expected, in most cases,
      the test error is slightly
      better than the results using $ 1500 $ decision stumps in
      Table \ref{TAB:QP1}; and no significant difference is observed.
      This verifies the capability of \adaboost-QP for selecting
      and combining relevant weak classifiers.



         \begin{table*}[t]
         \caption
         {
         Test results of \adaboost (AB) and \adaboost-QP (QP).
         All tests are run $ 10 $ times. The mean and standard
         deviation are reported. 
         \adaboost-QP outperforms \adaboost on $ 9 $
         datasets. 
         }
         \centering
         \begin{footnotesize}
         \begin{tabular}{l l l l l l}
         \hline\hline
dataset &  algorithm  & test error         
          & minimum margin   & average margin\\
\hline
{\bf  australian}
& AB
           &$0.153\pm0.034$      &$\bf -0.012\pm0.005$      &$0.082\pm0.006$\\ 
& QP
           &$\bf 0.13\pm0.038$          &$-0.227\pm0.081$      &$\bf 0.18\pm0.052$\\ 
\hline
{\bf  b-cancer}
& AB
           &$0.041\pm0.013$               &$\bf 0.048\pm0.009$       &$0.209\pm0.02$\\ 
& QP
           &$\bf 0.03\pm0.012$   
           &$-0.424\pm0.250$      &$\bf 0.523\pm0.237$\\ 
\hline
{\bf  diabetes}
& AB
           &$0.270\pm0.043$      &$\bf -0.038\pm0.007$       &$0.055\pm0.005$\\ 
& QP
           &$\bf 0.262\pm0.047$   
           &$-0.107\pm0.060$      &$\bf 0.075\pm0.031$\\ 
\hline
{\bf  fourclass}
& AB
           &$\bf 0.088\pm0.032$      &$\bf -0.045\pm0.012$       &$0.084\pm0.009$\\ 
& QP
           &$0.095\pm0.028$   &$-0.211\pm0.059$
           &$\bf 0.128\pm0.027$\\ 
\hline
{\bf  g-numer}
& AB
           &$0.283\pm0.033$   &$\bf -0.079\pm0.017$
           &$0.042\pm0.006$\\ 
& QP
           &$\bf 0.249\pm0.033$ 
           &$-0.151\pm0.058$      &$\bf 0.061\pm0.020$\\ 
\hline
{\bf  heart}
& AB
           &$0.210\pm0.032$      &$\bf 0.02\pm0.008$       &$0.104\pm0.013$\\ 
& QP
           &$\bf 0.190\pm0.058$  &$-0.117\pm0.066$
           &$\bf 0.146\pm0.059$\\ 
\hline
{\bf  ionosphere}
& AB
           &$\bf 0.121\pm0.044$  &$\bf 0.101\pm0.010$       &$0.165\pm0.012$\\ 
& QP
           &$0.139\pm0.055$     &$-0.035\pm0.112$
           &$\bf 0.184\pm0.063$\\ 
\hline
{\bf liver}
& AB
           &$0.321\pm0.040$      &$\bf -0.012\pm0.007$       &$0.055\pm0.005$\\ 
& QP
           &$\bf 0.314\pm0.060$  
           &$-0.107\pm0.044$      &$\bf 0.079\pm0.021$\\ 
\hline
{\bf  mushrooms}
& AB
           &$\bf 0\pm0$           &$\bf 0.102\pm0.001$       &$0.181\pm0.001$\\ 
& QP
           &$0.005\pm0.002$    &$-0.134\pm0.086$
           &$\bf 0.221\pm0.084$\\ 
\hline
{\bf  sonar}
& AB
           &$\bf 0.145\pm0.046$   &$\bf 0.156\pm0.008$       &$0.202\pm0.013$\\ 
& QP
           &$0.171\pm0.048$     &$0.056\pm0.066$
           &$\bf 0.220\pm0.045$\\ 
\hline
{\bf  splice}
& AB
           &$0.129\pm0.025$      &$\bf
           -0.009\pm0.008$       &$0.117\pm0.009$\\ 
& QP
           &$\bf 0.106\pm0.029$
           &$-0.21\pm0.037$      &$\bf 0.189\pm0.02$\\ 
\hline
{\bf  svmguide1}
& AB
           &$\bf 0.035\pm0.009$    &$\bf -0.010\pm0.008$       &$0.157\pm0.016$\\ 
& QP
           &$0.040\pm0.009$     &$-0.439\pm0.183$
           &$\bf 0.445\pm0.155$\\ 
\hline
{\bf  svmguide3}
& AB
           &$0.172\pm0.023$     &$\bf -0.011\pm0.009$       &$0.052\pm0.005$\\ 
& QP
           &$\bf 0.167\pm0.022$  
           &$-0.113\pm0.084$      &$\bf 0.085\pm0.038$\\ 
\hline
{\bf  w1a}
& AB
           &$0.041\pm0.014$       &$\bf -0.048\pm0.010$       &$0.084\pm0.005$\\ 
& QP
           &$\bf 0.029\pm0.009$ 
           &$-0.624\pm0.38$      &$\bf 0.577\pm0.363$\\ 
         \hline\hline
         \end{tabular}
         \end{footnotesize}
        \label{TAB:QP1}
         \end{table*}



         \begin{table*}[t]
         \caption
         {
         Test results of \adaboost-QP  on full sets of decision
         stumps.
         All tests are run $ 10 $ times. 
          }
         \centering
        \resizebox{0.999\textwidth}{!}
        {
         \begin{tabular}{l c c c c c c c c }
              \hline\hline
            %
            %
            %
            %
            %
%
         dataset &  \bf australian  & \bf b-cancer & \bf
         fourclass & \bf g-numer & \bf heart &
         \bf liver & \bf mushroom & \bf splice       \\
         test error & $0.131 \pm 0.041$ &  $0.03 \pm 0.011$ &
         $0.091\pm 0.02 $  & $ 0.243\pm 0.026 $ &  $0.188\pm 0.058$ 
         & $ 0.319\pm 0.05 $ & $0.003\pm 0.001$ & $0.097 \pm 0.02$  
         \\
         \hline\hline
        \end{tabular}
        }
        \label{TAB:QP2}
         \end{table*}

\subsection{\adaboost-CG}

      We run \adaboost and \adaboost-CG with decision stumps 
      on the datasets of \cite{LIBSVMdata2001}.
      %
      %
      %
      $ 70\% $ of
      examples are used for training; $ 15\% $ are used for test and
      the other $ 15\% $ are not used because we do not do
      cross-validation here. 
      The convergence threshold for \adaboost-CG 
      ($ \varepsilon $ in Algorithm \ref{alg:AdaCG}) is
      set to $10^{-5}$. 
%
%
      Another important parameter to tune is the regularization
      parameter $ T $.
      For the first experiment,
      we have set it to $ 1/\b{1}^\T \bw $ where 
      $ \bw $ is obtained by running \adaboost on the same data
      for $ 1000 $ iterations.
      Also for fair comparison, we have deliberately
      forced \adaboost-CG to run $ 1000 $ iterations even if the
      stopping criterion is met. 
      Both test and training results for \adaboost and \adaboost-CG
      are reported in Table~\ref{TAB:CG1} for a maximum number of
      iterations of $ 100 $, $500$ and $1000$.

      As expected, in terms of test error, no algorithm statistically 
      outperforms the other one, since they optimize the same cost
      function.
      As we can see, \adaboost does slightly better on $ 6 $ datasets.
      \adaboost-CG outperforms \adaboost on $ 7 $ datasets and on 
      {\em svmguide1}, both algorithms perform almost identically. 
      Therefore, empirically we conclude that in therms of
      generalization capability, \adaboost-CG is the same as the
      standard \adaboost.
    
      However, in terms of training error and convergence speed of the
      training procedure, there is significant difference between
      these two algorithms.       
      Looking at the right part of Table~\ref{TAB:CG1},        
      we see that the training error of \adaboost-CG is consistently
      better or no worse than \adaboost on {\em all} tested datasets.
      We have the following conclusions. 
      \begin{itemize}
         \item
            The convergence speed of \adaboost-CG is faster than
            \adaboost and in many cases, better training error can be
            achieved. 
            This is because \adaboost's coordinate descent nature is
            slow while \adaboost-CG is {\em totally
            corrective}\footnote{Like \lpboost, at each iteration
            \adaboost-CG updates the previous weak
            classifier weights $ \bw $.}. 
            This also means that with \adaboost-CG, we can use fewer
            weak classifiers to build a good strong classifier. 
            This is desirable for real-time applications like face
            detection 
            \cite{Viola2004Robust}, in which the testing speed is critical. 
         \item
            Our experiments confirm that a smaller training error does
            not necessarily lead to a smaller test error. This has
            been studied extensively in statistical learning theory. 
            It is observed that \adaboost sometimes suffers from
            overfitting and minimizing the exponential cost function
            of the margins does not solely determine test error. 
      \end{itemize}

      In the second experiment, we 
      run cross-validation to select the best value for the
      regularization parameter $ T $, same as in
      Section~\ref{sec:qp_exp}.
      Table \ref{TAB:CG_CV} reports the test errors on a subset of the
      datasets. Slightly better results are obtained compared with the
      results in Table \ref{TAB:CG1}, which uses $ T $ determined by \adaboost.

%
%
%
%
         \begin{table*}[ht]
         \caption
         {
               Test and training errors of \adaboost (AB) and
               \adaboost-CG (CG). All tests are run $ 5 $ times.
               The mean and standard deviation are reported. 
               Weak classifiers are decision stumps. 
         }
         \centering
         \resizebox{0.999\textwidth}{!}
         {
             \begin{tabular}{llllllll}
         \hline\hline
   dataset    & algorithm  & test error $100$  & test error $500$  &
   test
   error $1000$ & train error $100$      &  train error $500$      &  train
   error $1000$    \\
\hline
{\bf{australian}}
& AB
                             &$\bf 0.146\pm0.028$ &$\bf 0.165\pm0.018$
                             &$\bf 0.163\pm0.021$ &$0.091\pm0.013$ &$0.039\pm0.011$ &$0.013\pm0.009$  \\
& CG
                             &$0.177\pm0.025$ &$0.167\pm0.023$
                             &$0.167\pm0.023$ &$\bf 0.013\pm0.008$
                             &$\bf 0.011\pm0.007$ &$\bf 0.011\pm0.007$  \\ 

\hline

{\bf  {b-cancer}}
& AB
                             &$\bf 0.041\pm0.026$ &$\bf 0.045\pm0.030$
                             &$\bf 0.047\pm0.032$ &$0.008\pm0.006$ &$0\pm0$         &$0\pm0$  \\
& CG
                             &$0.049\pm0.033$ &$0.049\pm0.033$
                             &$0.049\pm0.033$ &$\bf 0\pm0$
                             &$ 0\pm0$         &$ 0\pm0$  \\ 
\hline
{\bf  {diabetes}}
& AB
                             &$\bf 0.254\pm0.024$ &$0.263\pm0.028$ &$0.257\pm0.041$ &$0.171\pm0.012$ &$0.120\pm0.007$ &$0.082\pm0.006$  \\
& CG
                             &$0.270\pm0.047$ &$\bf 0.254\pm0.026$
                             &$\bf 0.254\pm0.026$ &$\bf 0.083\pm0.008$
                             &$\bf 0.070\pm0.007$ &$\bf 0.070\pm0.007$  \\ 
\hline
{\bf  {fourclass}}
& AB
                             &$0.106\pm0.047$ &$0.097\pm0.034$ &$0.091\pm0.031$ &$0.072\pm0.023$ &$0.053\pm0.017$ &$0.046\pm0.017$  \\
& CG
                             &$\bf 0.082\pm0.031$ &$\bf 0.082\pm0.031$
                             &$\bf 0.082\pm0.031$ &$\bf 0.042\pm0.015$
                             &$\bf 0.042\pm0.015$ &$\bf 0.042\pm0.015$  \\ 
\hline
{\bf  {g-numer}}
& AB
                             &$0.279\pm0.043$ &$0.288\pm0.048$ &$0.297\pm0.051$ &$0.206\pm0.047$ &$0.167\pm0.072$ &$0.155\pm0.082$  \\
& CG
                             &$\bf 0.269\pm0.040$ &$\bf 0.262\pm0.045$
                             &$\bf 0.262\pm0.045$ &$\bf 0.142\pm0.077$
                             &$\bf 0.142\pm0.077$ &$\bf 0.142\pm0.077$  \\ 
\hline
{\bf  {heart}}
& AB
                             &$0.175\pm0.073$ &$0.175\pm0.088$ &$0.165\pm0.076$ &$0.049\pm0.022$ &$0\pm0$         &$0\pm0$  \\
& CG
                             &$\bf 0.165\pm0.072$ &$\bf 0.165\pm0.072$
                             &$\bf 0.165\pm0.072$ &$\bf 0\pm0$         &$0\pm0$         &$0\pm0$  \\ 
\hline
{\bf  {ionosphere}}
& AB
                             &$\bf 0.092\pm0.016$ &$\bf 0.104\pm0.017$
                             &$\bf 0.100\pm0.016$ &$0\pm0$         &$0\pm0$         &$0\pm0$  \\
& CG
                             &$0.131\pm0.034$ &$0.131\pm0.034$ &$0.131\pm0.034$ &$0\pm0$         &$0\pm0$         &$0\pm0$  \\ 
\hline
{\bf  {liver}}
& AB
                             &$0.288\pm0.101$ &$\bf 0.265\pm0.081$ &$\bf 0.281\pm0.062$ &$0.144\pm0.018$ &$0.063\pm0.015$ &$0.020\pm0.015$  \\
& CG
                             &$\bf 0.288\pm0.084$ &$0.288\pm0.084$
                             &$0.288\pm0.084$ &$\bf 0.017\pm0.012$
                             &$\bf 0.017\pm0.011$ &$\bf 0.017\pm0.011$  \\ 
\hline
{\bf  {mushrooms}}
& AB
                             &$0\pm0.001$     &$0\pm0.001$     &$0\pm0$         &$0\pm0$         &$0\pm0$         &$0\pm0$\\
& CG
                             &$\bf 0\pm0$         &$\bf 0\pm0$         &$0\pm0$         &$0\pm0$         &$0\pm0$         &$0\pm0$\\ 
\hline
{\bf  {sonar}}
& AB
                             &$\bf 0.206\pm0.087$ &$\bf 0.213\pm0.071$
                             &$\bf 0.206\pm0.059$ &$0\pm0$         &$0\pm0$         &$0\pm0$\\
& CG
                             &$0.232\pm0.053$ &$0.245\pm0.078$ &$0.245\pm0.078$  &$0\pm0$         &$0\pm0$         &$0\pm0$\\ 
\hline
{\bf  {splice}}
& AB
                             &$\bf 0.129\pm0.011$ &$\bf 0.143\pm0.026$
                             &$\bf 0.143\pm0.020$ &$0.053\pm0.003$ &$0.008\pm0.006$ &$0.001\pm0.001$  \\
& CG
                             &$0.161\pm0.033$ &$0.151\pm0.023$
                             &$0.151\pm0.023$ &$\bf 0.002\pm0.002$
                             &$\bf 0.001\pm0.002$ &$\bf 0.001\pm0.002$  \\ 
\hline
{\bf  {svmguide1}}
& AB
                             &$\bf 0.036\pm0.012$ &$\bf 0.034\pm0.008$ &$0.037\pm0.007$ &$0.022\pm0.002$ &$0.009\pm0.002$ &$0.002\pm0.001$  \\
& CG
                             &$0.037\pm0.007$ &$0.037\pm0.007$
                             &$0.037\pm0.007$ &$\bf 0.001\pm0.001$
                             &$\bf 0\pm0.001$     &$\bf 0\pm0.001$  \\ 
\hline
{\bf  {svmguide3}}
& AB
                             &$0.184\pm0.037$ &$0.183\pm0.044$ &$0.182\pm0.031$ &$0.112\pm0.009$ &$0.037\pm0.004$ &$0.009\pm0.003$  \\
& CG
                             &$\bf 0.184\pm0.026$ &$\bf 0.171\pm0.023$
                             &$\bf 0.171\pm0.023$ &$\bf 0.033\pm0.012$
                             &$\bf 0.023\pm0.016$ &$\bf 0.023\pm0.016$  \\ 
\hline
{\bf  {w1a}}
& AB
                             &$0.051\pm0.009$ &$0.038\pm0.005$ &$0.036\pm0.004$ &$0.045\pm0.008$ &$0.028\pm0.005$ &$0.025\pm0.005$  \\
& CG
                             &$\bf 0.018\pm0.001$ &$\bf 0.018\pm0.001$
                             &$\bf 0.018\pm0.001$ &$\bf 0.010\pm0.004$
                             &$\bf 0.010\pm0.004$ &$\bf 0.010\pm0.004$  \\ 
         \hline\hline
         \end{tabular}
         }
         \label{TAB:CG1}
         \end{table*}

\begin{figure*}[t!]
   \begin{center}
          \includegraphics[width=0.32\textwidth]{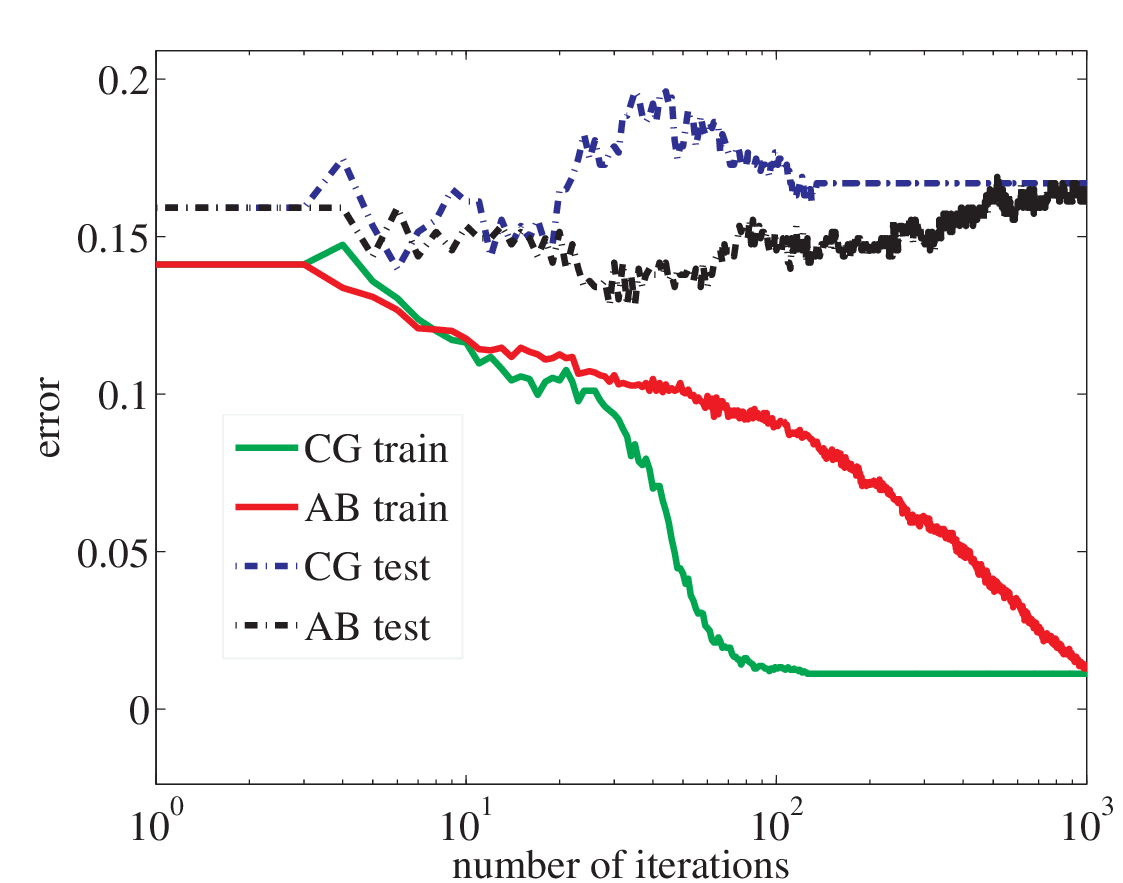}
          \includegraphics[width=0.32\textwidth]{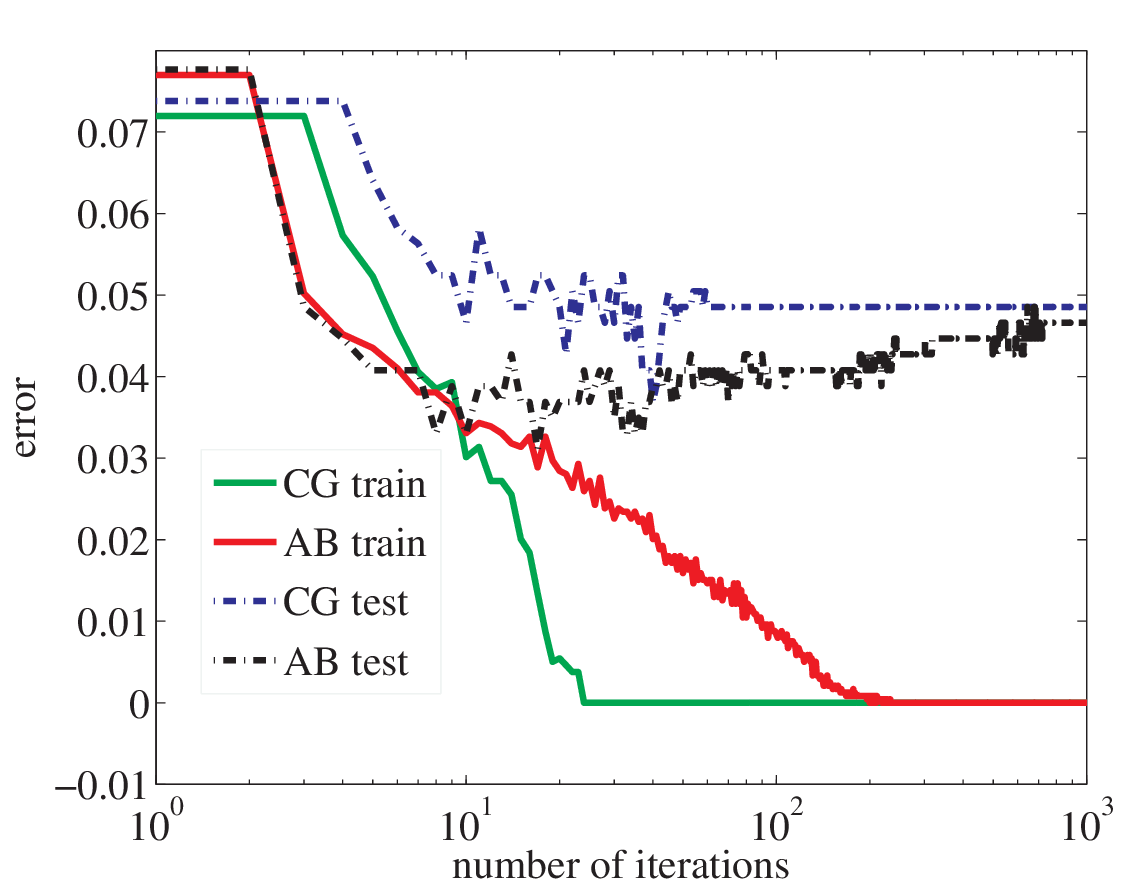}
          \includegraphics[width=0.32\textwidth]{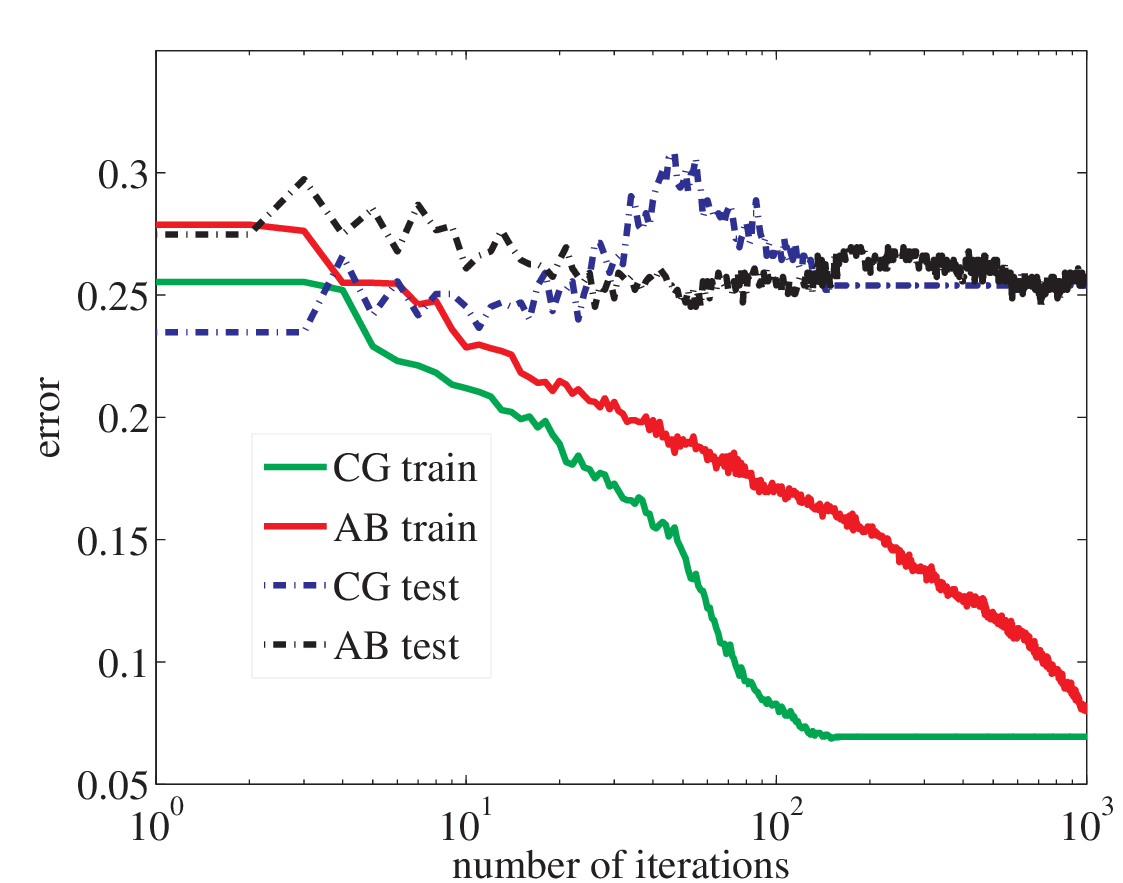}  
          \includegraphics[width=0.32\textwidth]{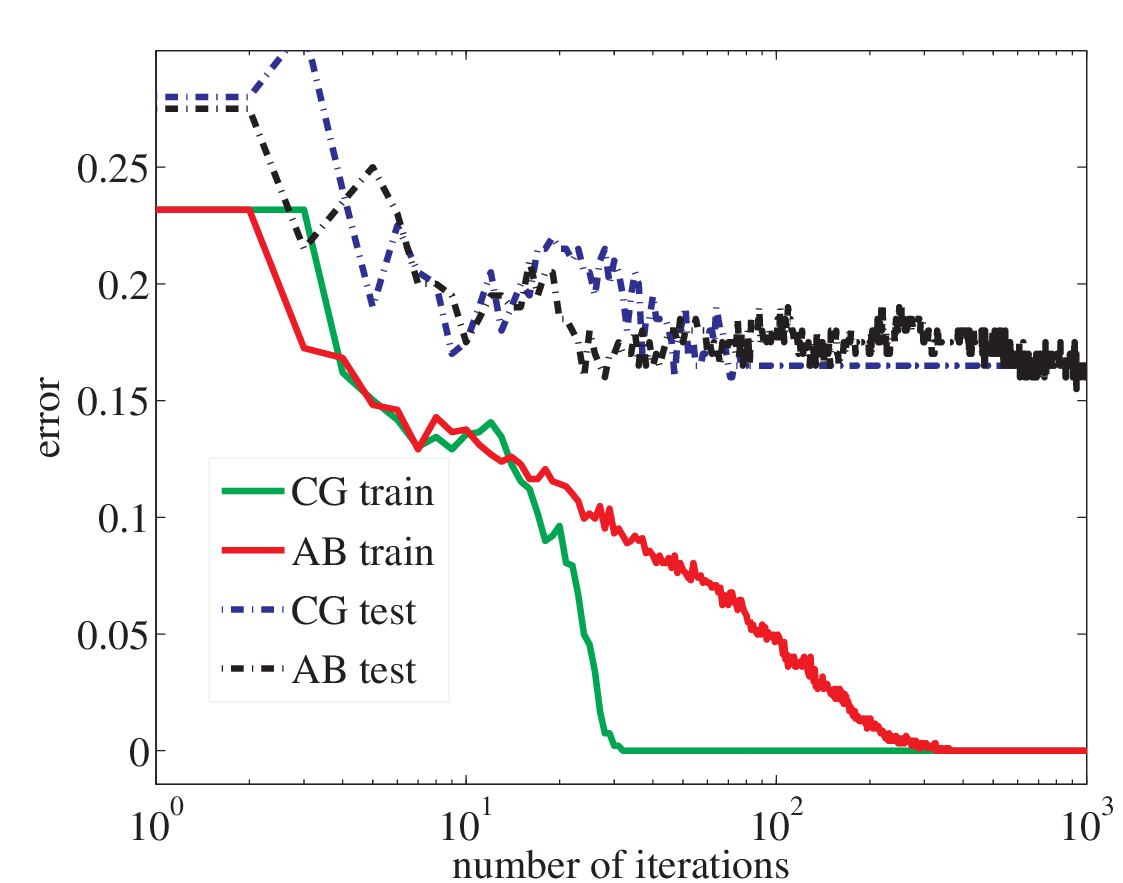}
          \includegraphics[width=0.32\textwidth]{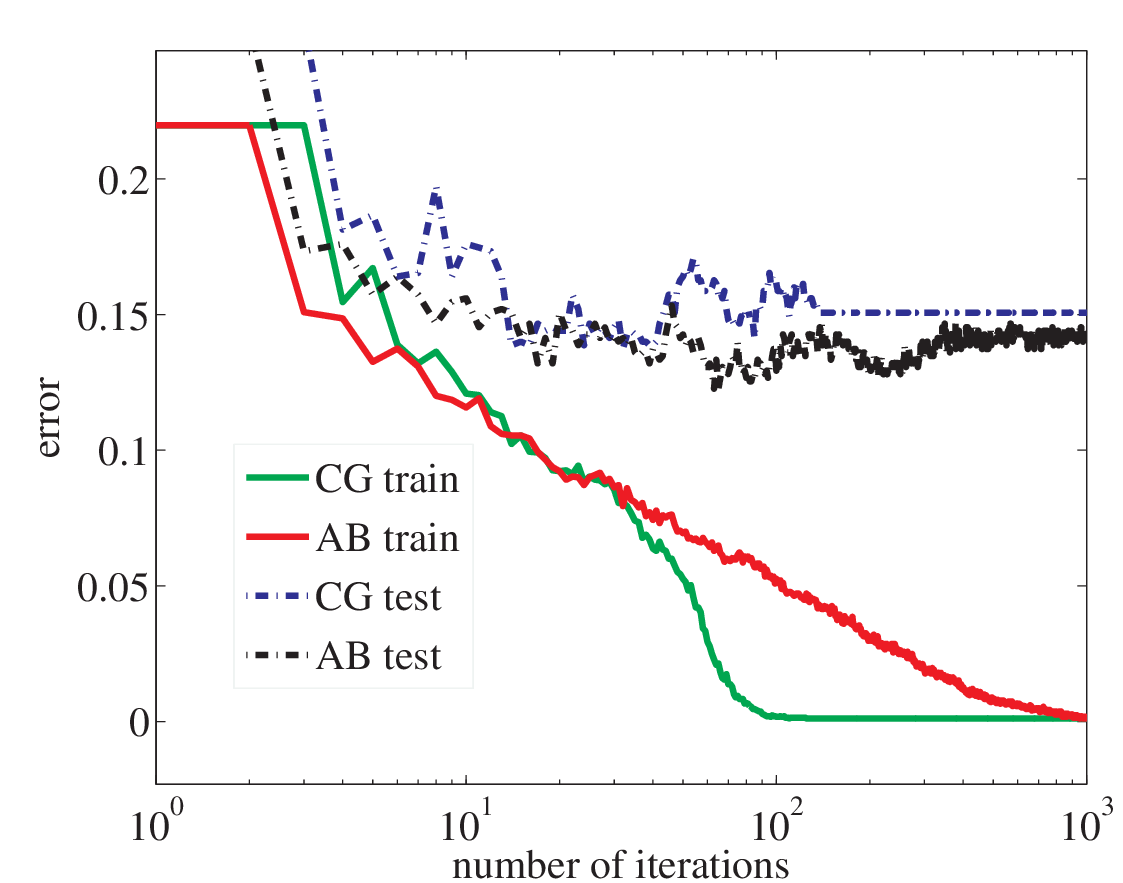}
          \includegraphics[width=0.32\textwidth]{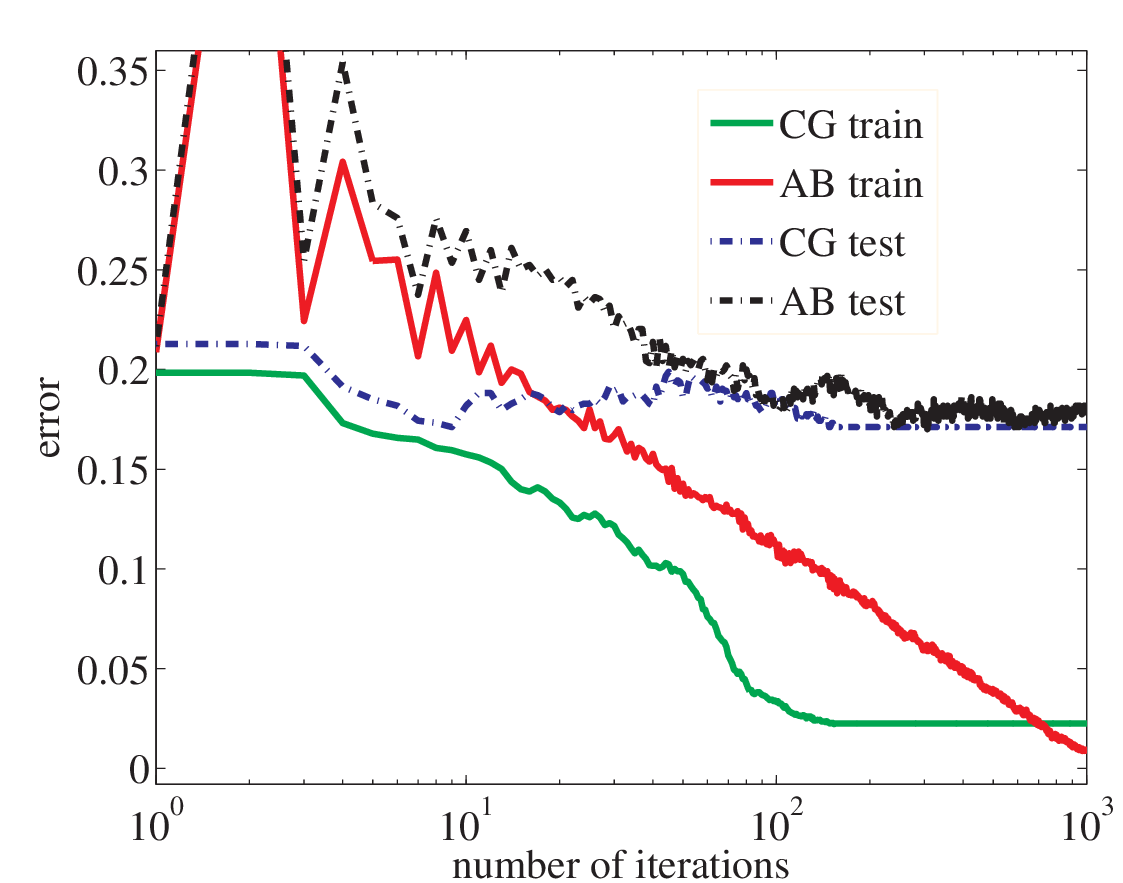}    
   \end{center}
   \caption{Test error and training error of \adaboost, \adaboost-CG
   for {\em australian, breast-cancer, diabetes, heart, spline} and 
   {\em svmguide3} datasets.
   These convergence curves correspond to the results in
   Table~\ref{TAB:CG1}. 
   The $x$-axis is on a logarithmic scale  
   for easier
   comparison. 
   }
   \label{fig:cg_err}
\end{figure*}
      

         \begin{table*}[t]
         \caption
         {
         Test error of \adaboost-CG with decision
         stumps, using cross-validation to select the optimal $ T
         $.
         All tests are run $ 5 $ times. 
          }
         \centering
          \resizebox{1\textwidth}{!}
        {
         \begin{tabular}{l c c c c c c c c }
              \hline\hline
         dataset &  \bf australian  & \bf b-cancer & \bf
         diabetes & \bf fourclass & \bf heart &
         \bf ionosphere & \bf sonar    & \bf splice   \\
         test error & $0.146 \pm 0.027$ 
                    & $0.033 \pm 0.033$ &
                      $0.266 \pm 0.036 $  & 
                      $ 0.086\pm 0.027 $ &  
                      $0.17\pm 0.082$ 
                    & $ 0.115\pm 0.024 $ 
                    & $0.2\pm 0.035$   
                    & $0.135 \pm 0.015 $
         \\
         \hline\hline
         \end{tabular}
        }
        \label{TAB:CG_CV}
         \end{table*}

      We also use LDA as weak classifiers to compare the
      classification performance of \adaboost and \adaboost-CG.
      The parameter $ T $ of \adaboost-CG
      is determined by cross-validation from $ 
      \{ $$ \frac{1}{2},$ 
         $ \frac{1}{5},$  
         $ \frac{1}{8}, $ 
         $ \frac{1}{10}, $ 
         $ \frac{1}{12},$ 
         $ \frac{1}{15},$ 
         $ \frac{1}{20},$ 
         $ \frac{1}{30},$ 
         $ \frac{1}{40},$ 
         $ \frac{1}{50},$ 
         $ \frac{1}{70},$ 
         $ \frac{1}{90},$ 
         $ \frac{1}{100},$ 
         $ \frac{1}{120},$ 
         $ \frac{1}{150}$$\}
      $.
      For \adaboost the smallest test error from $100$, $500$ and
      $1000$
      runs is reported. We show the results in Table
      \ref{TAB:CG_CV_LDA}. 
      As we can see, the test error is slightly better than with
      decision stumps for both \adaboost and \adaboost-CG.
      Again, \adaboost and \adaboost-CG's performances are very
      similar.

      In order to show that statistically there are no difference
      between \adaboost-CG and \adaboost, the McNemar test
      \cite{Dietterich1998Approximate} 
      with the significance level of $0.05$ is
      conducted. McNemar's test is based on a $ \chi^2$ test
      \cite{Dietterich1998Approximate}. If the quantity of the 
      $ \chi^2$ test is not greater than $ \chi^2_{1,0.95} = 3.841459$,
      we can think of that the two tested classifiers have 
      {\em no statistical difference} in
      terms of classification capability. 
      On the $ 8 $ datasets with decision stumps and LDA
      (Tables~\ref{TAB:CG_CV} and \ref{TAB:CG_CV_LDA}),
      in all cases ($ 5 $ runs per dataset), the results of $ \chi^2 $
      test are smaller than $ \chi^2_{1,0.95} $.
      Consequently, 
      we can conclude that indeed \adaboost-CG performs very similarly
      to \adaboost for classification.


         \begin{table*}[t]
         \caption
         {
         Test error of \adaboost (AB) and  \adaboost-CG (CG) with LDA
         as weak classifiers, using cross-validation to select the
         optimal $ T $.  All tests are run $ 5 $ times. 
         }
         \centering
         \resizebox{1\textwidth}{!}
         {
         \begin{tabular}{l c c c c c c c c }
              \hline\hline
         dataset &  \bf australian  & \bf b-cancer & \bf
         diabetes & \bf fourclass & \bf heart &
         \bf ionosphere & \bf sonar    & \bf splice   \\
         AB         & $0.150 \pm 0.044$ 
                    & $0.029\pm0.014$  &
                      $0.259\pm0.021$  & 
                      $0.003\pm0.004$  &  
                      $0.16\pm 0.055$ 
                    & $ 0.108\pm 0.060  $ 
                    & $ 0.297\pm 0.080  $   
                    & $ 0.215 \pm 0.027 $
         \\
         CG & $0.151\pm0.053$
            & $0.035\pm0.016$
            & $0.249\pm0.038$
            & $0.022\pm0.015$
            & $0.185\pm0.038$
            & $0.104\pm0.062$
            & $0.258\pm0.085$
            & $0.235\pm0.035$
         \\
         \hline\hline
         \end{tabular}
        }
        \label{TAB:CG_CV_LDA}
         \end{table*}

      To examine the effect of parameter $ T $, 
      we run more experiments with various $ T $ on the {\em banana}
      dataset (2D artificial data)
      that was used in
      \cite{Ratsch2001Soft}.
      We still use decision stumps. The maximum iteration is set to $
      400 $. All runs stop earlier than $ 100 $ iterations. 
      Table \ref{TAB:cg-banana} reports the results. 
      Indeed, the training error depends on $ T $.
       $ T $ also has influence on the
      convergence speed. 
      But, in a wide range of  $ T $, the test error does not change
      significantly.
      We do not have a sophisticated technique to tune $ T $. As
      mentioned, the sum of $\bw $ from a run of \adaboost 
      can serve as a heuristic.


%
%
%
%
         \begin{table}[h!]
         \caption
         {\adaboost-CG on {\em banana} dataset with decision stumps
         and LDA as weak classifiers. 
         Experiments are run $ 50 $ times.  
         }
         \centering
         \resizebox{0.495\textwidth}{!}
         {
         \begin{tabular}{l l l l l}
         \hline\hline
$ \frac{1}{T} $  &  test (stumps) & train (stumps)  & test (LDA)          & train (LDA)              \\
\hline 
$ 20 $   & $0.298 \pm 0.018 $  & $ 0.150  \pm 0.019  $   & $0.134 \pm 0.012 $  & $ 0.032  \pm 0.007  $   \\
$ 40 $   & $0.309 \pm 0.019 $  & $ 0.101  \pm 0.015  $   & $0.135 \pm 0.008 $  & $ 0.001  \pm 0.002  $   \\
$80  $   & $0.313 \pm 0.019 $  & $ 0.033  \pm 0.011  $   & $0.136 \pm 0.007 $  & $ 0  \pm 0  $ \\ 
         \hline\hline
         \end{tabular}
         }
         \label{TAB:cg-banana}
         \end{table}


         \begin{table*}[t!]
         \caption
         {
         Test error of LogitBoost-CG with decision
         stumps, using cross-validation to select the optimal $ T
         $.
         All tests are run $ 5 $ times. 
          }
         \centering
          \resizebox{1\textwidth}{!}
        {
         \begin{tabular}{l c c c c c c c c }
              \hline\hline
         dataset &  \bf australian  & \bf b-cancer & \bf
         diabetes & \bf fourclass & \bf heart &
         \bf ionosphere & \bf sonar    & \bf splice   \\
         test error & $0.13 \pm 0.043 $ 
                    & $0.039 \pm 0.012 $ &
                      $0.238 \pm 0.057 $  & 
                      $ 0.071\pm 0.034 $ &  
                      $0.14\pm 0.095$ 
                    & $ 0.2\pm 0.069 $ 
                    & $0.169\pm 0.05$   
                    & $0.104 \pm 0.021 $
         \\
         \hline\hline
         \end{tabular}
        }
        \label{TAB:LOGIT_CV}
         \end{table*}

      Now let us take a close look at the convergence behavior of
      \adaboost-CG. 
      \fig~\ref{fig:cg_err} shows the test and training error of
      \adaboost and \adaboost-CG for $ 6 $ datasets.   
      We see that \adaboost-CG converges much faster than \adaboost
      in terms of number of iterations.
      On most tested datasets, \adaboost-CG is around $10$ times
      faster than \adaboost.  
      The test error for these two methods are very close upon
      convergence. 
      In some datasets such as {\em australian} and {\em
      breast-cancer}
      we observe over-fitting for \adaboost.

       \subsection{\logitboost-CG}
       \label{sec:exp-logit}
        
       We have also run \logitboost-CG on the same datasets. 
       All the settings are the same as in the case of \adaboost-CG. 
       The weak classifiers are decision stumps. 
       Table~\ref{TAB:LOGIT_CV} reports the experiment results.
       Compared to Table~\ref{TAB:CG_CV}, very similar results have been
       observed. No one achieves better results over the other one
       on all the datasets.


\section{Discussion and Conclusion}

      In this paper, we have shown  that  the Lagrange dual problems
      of \adaboost, \logitboost and soft-margin \lpboost with
      generalized hinge loss are all entropy regularized \lpboost. We
      both theoretically and empirically demonstrate that the success
      of \adaboost relies on maintaining a better margin distribution.
      Based on the dual formulation, a general column generation based
      optimization framework is proposed. This optimization framework
      can be applied to solve all the boosting algorithms with
      various loss functions mentioned in this paper.  
      Experiments with exponential loss show that the
      classification performance of \adaboost-CG is statistically
      almost identical to the standard stage-wise \adaboost on real
      datasets. 
      In fact, since both algorithms optimize the same cost function,
      we would be surprised to see a significant different in their
      generalization error.
      The main advantage of the proposed algorithms is significantly
      faster convergence speed.

      Compared with the conventional \adaboost, a drawback of
      \adaboost-CG is the introduction of a parameter, same as in
      \lpboost. While one can argue that \adaboost implicitly
      determines this same parameter by selecting how many iterations
      to run, the stopping criterion is nested and thus efficient to
      learn. In the case of \adaboost-CG, it is not clear how to
      efficiently learn this parameter. Currently, one has to run the
      training procedure multiple times for cross validation.

      With the optimization framework established here, some
      issues on boosting that are previously unclear may become
      obvious now. For example, for designing cost-sensitive boosting
      or boosting on uneven datasets, one can simply modify the primal
      cost function \eqref{EQ:2} to have a weighted cost function
      \cite{Leskovec2003LPBoost}.  The training procedure follows
      \adaboost-CG.

      To summarize, the convex duality of boosting algorithms
      presented in this work generalizes the convex duality in
      \lpboost. We have shown some interesting properties that the
      derived dual formation possesses. The duality also leads to new
      efficient learning algorithms. The duality provides
      useful insights on boosting that may lack in existing
      interpretations \cite{Schapire1998Margin, Friedman2000Additive}.

      In the future, we want to extend our work to boosting with
      non-convex loss functions such as BrownBoost
      \cite{Freund2001Adaptive}. Also it should be straightforward to
      optimize boosting for regression using column generation. 
      We are currently exploring the application of \adaboost-CG to
      efficient object detection due to its faster convergence, which
      is more promising for feature selection \cite{Viola2004Robust}.
 %
 %


\appendices

\section{Description of datasets}


         \begin{table*}[t]
         \caption
         {
         Description of the datasets.
         Except {\em mushrooms},
         {\em svmguide1}, {\em svmguide3} and {\em w1a},
         all the other datasets have been scaled to
         $ [-1,1] $. 
         }
         \centering
         \begin{small}
         \begin{tabular}{l l l||l l l}

         \hline\hline
dataset  & \# examples  & \# features      & dataset     & \# examples & \# features\\
\hline\footnotesize

{\bf {australian}}     & 690       & 14            &{\bf {liver-disorders}}          & 345        & 6\\
{\bf {breast-cancer}} & 683       & 10            &{\bf {mushrooms}}       & 8124       & 112\\
{\bf {diabetes}}       & 768       & 8             &{\bf {sonar}}          & 208        & 60\\
{\bf {fourclass}}      & 862       & 2             &{\bf {splice}}         & 1000       & 60\\
{\bf {german-numer}}         & 1000       & 24            &{\bf {svmguide1}}           & 3089       & 4\\
{\bf {heart}}          & 270       & 13            &{\bf {svmguide3}}           & 1243       & 22\\
{\bf {ionosphere}}            & 351       & 34            &{\bf {w1a}}            & 2477       & 300\\

         \hline\hline

         \end{tabular}
         \end{small}
         \label{TAB:dataset1}
         \end{table*}

 Table~\ref{TAB:dataset1} provides a description of the datasets we
 have used in the experiments.

%
%

 \section*{Acknowledgments}

        NICTA is funded by the Australian Government as represented by
        the Department of Broadband, Communications and the Digital
        Economy and the Australian Research Council through the ICT
        Center of Excellence program.

        The authors thank Sebastian Nowozin for helpful discussions.

 \footnotesize
 \bibliographystyle{ieee}

\end{document}